\documentclass{article}
\usepackage{kr}

\usepackage{times}
\usepackage{soul}
\usepackage{url}
\usepackage[hidelinks]{hyperref}
\usepackage[utf8]{inputenc}
\usepackage[small]{caption}
\usepackage{graphicx}
\usepackage{amsmath}
\usepackage{amsthm}
\usepackage{booktabs}
\usepackage{algorithm}
\usepackage{algorithmic}
\usepackage{amssymb}
\usepackage{amsthm}

\usepackage[dvipsnames]{xcolor}

\definecolor{color1}{RGB}{234, 32, 39} 
\definecolor{color2}{RGB}{87, 88, 187}
\definecolor{color3}{RGB}{247, 159, 31} 
\definecolor{color4}{RGB}{0, 148, 50}

\definecolor{Myblue}{RGB}{0, 184, 148}

\definecolor{centuryred}{RGB}{246, 49, 46}

\usepackage{thmtools}
\usepackage{thm-restate}
\usepackage{cleveref}

\declaretheoremstyle[
    headfont=\normalfont\bfseries, 
    bodyfont = \normalfont,
    postheadspace=1em, 
    qed=$\square$, 
    headpunct={:}]{mystyle} 
\declaretheorem[name={Proof}, style=mystyle, unnumbered]{myproof}

\newtheorem{example}{Example}
\newtheorem{definition}{Definition}
\newtheorem*{lemma*}{Lemma}

\usepackage{tikz}
\usetikzlibrary{calc}
\usepackage{booktabs}
\usetikzlibrary{arrows}
\usetikzlibrary{arrows.meta}
\usepackage{pgf}
\usetikzlibrary{arrows,automata}
\usetikzlibrary{positioning}
\usetikzlibrary{fit,backgrounds}

\newcommand{\tikzmark}[1]{\tikz[overlay,remember picture] \node (#1) {};}
\newcommand{\DrawBox}[3][]{%
    \tikz[overlay,remember picture]{
    \draw[black,#1]
      ($(#2)+(-0.5em,2ex)$) rectangle
      ($(#3)+(0.75em,-0.75ex)$);}
}

\tikzset{
    state/.style={
    text width=30mm,
           rectangle,
           rounded corners,
           draw=black, very thick,
           minimum height=2em,
           inner sep=0pt,
           align=center,
           },
}

\makeatletter \g@addto@macro\@floatboxreset\centering \makeatother %center all tikz pictures

\makeatletter
\newcommand\mathcircled[1]{%
  \mathpalette\@mathcircled{#1}%
}
\newcommand\@mathcircled[2]{%
  \tikz[baseline=(math.base)] \node[draw,circle,inner sep=1pt] (math) {$\m@th#1#2$};%
}
\makeatother

\newcommand{\mods}[1]{[\![#1]\!]}
\newcommand{\bel}[1]{[#1]}
\newcommand{\cbel}[1]{[#1]_{\mathrm{c}}}
\newcommand{\contract}{\div} 

\newcommand{\STQ}{\oplus_{\mathrm{STQ}}}

\newcommand{\RT}{$(\mathrm{RT})$}
\newcommand{\DERevR}{$(\mathrm{DE}^{\ast}_{\scriptscriptstyle \preccurlyeq})$} %Doxastic equivalence

\newcommand{\SRevR}{$(\mathrm{S}^{\ast}_{\scriptscriptstyle \preccurlyeq})$}
\newcommand{\SRevS}{$(\mathrm{S}^{\ast})$}

\newcommand{\BetaRevR}[1]{$(\beta{#1}^{\ast}_{\scriptscriptstyle \preccurlyeq})$}

\newcommand{\CRevR}[1]{$(\mathrm{C}{#1}^{\ast}_{\scriptscriptstyle \preccurlyeq})$}

\newcommand{\KRev}[1]{$(\mathrm{K}{#1}^{\ast})$}

\newcommand{\Cn}{\mathrm{Cn}}
\newcommand{\C}{\mathrm{C}}
\newcommand{\CRat}{\mathrm{C_{rat}}}

\newcommand{\astN}{\ast_{\mathrm{N}}}
\newcommand{\astR}{\ast_{\mathrm{R}}}
\newcommand{\astL}{\ast_{\mathrm{L}}}

\newcommand{\plusBG}{+_{\mathrm{BG}}}
\newcommand{\contractBG}{\contract_{\mathrm{BG}}}

\newcommand{\pprincRevR}[1]{$(\mathrm{Ret}{#1}^{\ast}_{\preccurlyeq})$}

\newcommand{\pprincRev}[1]{$(\mathrm{Ret}{#1})$}

\newcommand{\KIRevR}[1]{$(\mathrm{KI}{#1}^{\ast}_{\preccurlyeq})$}

\usepackage{stmaryrd}
\usepackage{trimclip}

\makeatletter
\DeclareRobustCommand{\shortTo}{%
  \mathrel{\mathpalette\short@To\relax}%
}

\newcommand{\short@To}[2]{%
  \mkern2mu
  \clipbox{{.4\width} 0 0 0}{$\m@th#1\vphantom{+}{\Rightarrow}$}%
  }
\makeatother

\newcommand{\shortto}{\supset}

\usepackage{enumitem}
\setlist[itemize,1]{leftmargin=3em}

\title{Revision by Conditionals: \\ From Hook to Arrow}

\author{%
Jake Chandler$^1$\and
Richard Booth$^2$ \\
\affiliations
$^1$La Trobe University\\
$^2$Cardiff University\\
\emails
jacob.chandler@latrobe.edu.au,
boothr2@cardiff.ac.uk
}

\begin{document}

\maketitle

%\begin{abstract}
%The belief revision literature has largely focussed on the issue of how to revise one's beliefs in the light of information regarding matters of fact. Here we turn to an important but comparatively neglected issue: How to model agents capable of acquiring information regarding which rules of inference (`Ramsey Test conditionals') they ought to use in reasoning about these facts. Our approach to this second question of so-called `conditional revision'  is distinctive insofar as it abstracts from the controversial details of how to address the first. We introduce a `plug and play' method for uniquely extending any iterated belief revision operator to the conditional case. The flexibility of our approach is achieved by having the result of a conditional revision by a Ramsey Test conditional (`arrow') determined by that of a plain revision by its corresponding material conditional (`hook'). It is shown to satisfy a number of new constraints that are of independent interest.
%\end{abstract}

\begin{abstract}
The belief revision literature has largely focussed on the issue of how to revise one's beliefs in the light of information regarding matters of fact. Here we turn to an important but comparatively neglected issue: How might one extend a revision operator to handle {\em conditionals} as input? Our approach to this question of `conditional revision'  is distinctive insofar as it abstracts from the controversial details of how to revise by factual sentences. We introduce a `plug and play' method for uniquely extending any iterated belief revision operator to the conditional case. The flexibility of our approach is achieved by having the result of a conditional revision by a Ramsey Test conditional (`arrow') determined by that of a plain revision by its corresponding material conditional (`hook'). It is shown to satisfy a number of new constraints that are of independent interest.
\end{abstract}

\section{Introduction}

The past three decades have witnessed the development of a substantial, if inconclusive, body of work devoted to the issue of {\em belief revision}, namely
\begin{itemize}

\item[(A)]  determining the impact of a local change in belief on both (i) the remainder of one’s prior beliefs and (ii) one’s prior {\em conditional beliefs} (`Ramsey Test conditionals'). 
 
\end{itemize}
 Surprisingly, however, very little has been done to this date on  the question of {\em conditional belief revision}, that is
 \begin{itemize}

\item[(B)]   determining the impact of a local change in conditional beliefs on both (i) and (ii). 
 
\end{itemize}
Furthermore, nearly all of the few proposals to tackle  issue (B), namely \cite{Hansson1992-HANIDO-3},   \cite{DBLP:conf/aaai/BoutilierG93}, and  \cite{DBLP:conf/ecai/NayakPFP96}, have typically rested on somewhat contentious assumptions about how to approach (A). (A noteworthy  exception to this \cite{DBLP:conf/ijcai/Kern-Isberner99}, who introduced a number of plausible general postulates governing revision by conditionals whose impact on revision simpliciter remains fairly modest. More on these below.)

In this paper, we consider the prospects of providing a ‘plug and play’ solution to issue (B) that is  independent of the details of how to address (A). Its remainder is organised as follows. First, in Section \ref{s:Revision}, we present some standard background on problem (A), introducing along the way the well-known  notion of a Ramsey Test conditional or again conditional belief. In Section \ref{s:CRevision}, we outline and discuss our proposal regarding (B). Subsection \ref{ss:Aproposal}  presents the basic idea, according to which computing the result of a revision by a Ramsey Test conditional can be derived by minimal modification, under constraints, of the outcome of a revision by its corresponding material conditional. Our key technical contribution is presented in Subsection \ref{ss:RResult}, where we prove that this minimal change under constraints can be achieved by means of a simple and familiar transformation. Subsection \ref{ss:AltChar} outlines some interesting general properties of the proposal. These strengthen, in a plausible manner, the aforementioned constraints presented in \cite{DBLP:conf/ijcai/Kern-Isberner99} and are of independent interest. Subsection \ref{ss:Elementary} considers the upshot of pairing our proposal regarding (B) with some well-known suggestions regarding how to tackle (A). Finally, in Section \ref{s:RResearch}, we  compare the suggestion made with existing work on the topic noting some important shortcomings of the latter. We close the paper in Section \ref{s:CComments} with a number of questions for future research. The proofs of the various propositions and theorems have been relegated to a technical appendix.

%The proofs of the various propositions and theorems have been relegated to a technical appendix.

%=====================================================

\section{Revision} 

\label{s:Revision}

%=====================================================

The beliefs of an agent are represented by a {\em belief state}. Such states will be denoted by upper case Greek letters $\Psi, \Theta,\ldots$. We denote by $\mathbb{S}$ the set of all such states. Each state determines a {\em belief set}, a consistent and deductively closed set of sentences, drawn from a finitely generated propositional, truth-functional language $L$, equipped with the standard connectives $\shortto$, $\wedge$, $\vee$, and $\neg$.
We denote the belief set associated with state $\Psi$ by $\bel{\Psi}$. Logical equivalence is denoted by $\equiv$ and the set of classical logical consequences of $\Gamma\subseteq L$ by $\mathrm{Cn}(\Gamma)$, with $\top$ denoting an arbitrary propositional tautology. The set of propositional worlds will be denoted by $W$ and the set of models of a given sentence $A$ by $\mods{A}$. 

The operation of {\em revision} $\ast$ returns the posterior state $\Psi \ast A$ that results from an adjustment of $\Psi$ to accommodate the inclusion of the consistent input $A$ in its associated belief set, in such a way as to maintain consistency of the resulting belief set. 
 
The beliefs resulting from single revisions are conveniently representable by a {\em conditional belief  set} $\cbel{\Psi}$, which can be viewed as encoding the agent's rules of inference over $L$ in state $\Psi$.
It is defined via the Ramsey Test:
\begin{tabbing}
\=BLAH:\=\kill
\>  \RT \>  For all $A, B \in L$, $A \shortTo B \in \cbel{\Psi}$ iff $B \in \bel{\Psi \ast A}$ \\[-0.25em]
\end{tabbing} 
\vspace{-0.5em}

\noindent We shall call $L_c$ the minimal extension of $L$ that additionally includes all sentences of the form $A \shortTo B$, with $A, B \in L$. We shall call sentences of the form $A\shortTo B$ `conditionals' and sentences of the form $A\shortto B$ `material conditionals'. We shall say that a sentence of the form $A\shortTo B$ is consistent just in case $A\wedge B$ is consistent (later in the paper, we shall explicitly disallow revisions by inconsistent conditionals). 

Conditional belief sets are constrained by the AGM postulates of \cite{alchourron1985logic,darwiche1997logic} (henceforth `AGM'). Given these, $\cbel{\Psi}$ corresponds to a  consistency-preserving {\em rational consequence relation}, in the sense of \cite{lehmann1992does}. Equivalently, it is representable by a  {\em total preorder (TPO)}  $\preccurlyeq_{\Psi}$ of worlds, such that $A\shortTo B\in \cbel{\Psi}$ iff $\min(\preccurlyeq_{\Psi}, \llbracket A\rrbracket)\subseteq \llbracket B\rrbracket$ \cite{grove1988two,katsuno1991propositional}. Note that $A\in\bel{\Psi}$ iff $\top\shortTo A\in\cbel{\Psi}$ or equivalently iff $\min(\preccurlyeq_{\Psi}, W)\subseteq \mods{A}$.

%We denote by $\TPO{W}$ the set of all TPOs over $W$.

Following convention, we shall call principles presented in terms of belief sets `syntactic', and call `semantic' those principles couched in terms of TPOs, denoting the latter by subscripting the corresponding syntactic principle with `$\preccurlyeq$'.
Due to space considerations and for ease of exposition, we will largely restrict our focus to a  semantic perspective on our problem of interest.

The AGM postulates do not entail 
% \jake{but they did!} 
 that one's conditional beliefs are determined by one's beliefs---in the sense that, if $\bel{\Psi}=\bel{\Theta}$, then  $\cbel{\Psi}=\cbel{\Theta}$---and there is widespread consensus that such determination would be unduly restrictive, with  \cite{Hansson1992-HANIDO-3} providing supporting arguments. A fortiori, one should not identify conditional beliefs with beliefs in the corresponding material conditional. That said, there does remain a connection between $A \shortTo B \in \cbel{\Psi}$  and $A \shortto B \in \bel{\Psi}$. The following is well known: 

\begin{restatable}{prop}{RamseyMat}
\label{RamseyMat}
Given AGM, (a) if $A \shortTo B \in \cbel{\Psi}$, then $A \shortto B \in \bel{\Psi}$, but (b) the converse does not hold.
\end{restatable}

\noindent Indeed, (a) is simply equivalent, given \RT, to the AGM postulate of Inclusion, according to which $\bel{\Psi\ast A}\subseteq \Cn(\bel{\Psi}\cup\{A\})$. This suggests the following catchline: 
\begin{itemize}
\item[] `Conditional beliefs are beliefs in material conditionals plus' 
\end{itemize}
That is, conditional beliefs are beliefs in material conditionals that satisfy certain additional constraints.

Regarding the conditional beliefs resulting from single revisions, i.e.~the beliefs resulting from sequences of two revisions, we assume an `irrelevance of syntax' property, which, in its semantic form, is given by:
\begin{tabbing}
\=BLAHI: \=\kill
\>  (Eq$^*_\preccurlyeq$) \> If $A\equiv B$, then $\preccurlyeq_{\Psi\ast A}=\preccurlyeq_{\Psi\ast B}$\\[-0.25em]
\end{tabbing} 
\vspace{-1em}
\noindent Given this principle, we take the liberty to abuse both language and notation and occasionally speak of revision by a set of worlds $S$ rather than by an arbitrary sentence whose set of models is given by $S$.

The DP postulates of  \cite{darwiche1997logic} provide widely endorsed further constraints. We simply give them here in their semantic form:

\begin{tabbing}
\=BLAHHH:\=\kill
\> \CRevR{1}  \> If $x,y \in \mods{A}$ then $x \preccurlyeq_{\Psi\ast A} y$ iff  $x \preccurlyeq_{\Psi} y$\\ [0.1cm]
\> \CRevR{2} \> If $x,y \in \mods{\neg A}$ then $x \preccurlyeq_{\Psi\ast A} y$ iff  $x \preccurlyeq_{\Psi} y$\\ [0.1cm]
\> \CRevR{3} \> If $x \in \mods{A}$, $y \in \mods{\neg A}$ and $x \prec_{\Psi} y$, then \\
\> \> $x \prec_{\Psi\ast A} y$ \\ [0.1cm]
\> \CRevR{4} \> If $x \in \mods{A}$, $y \in \mods{\neg A}$ and $x \preccurlyeq_{\Psi} y$, then\\
\> \>  $x \preccurlyeq_{\Psi\ast A} y$ \\[-0.25em]
\end{tabbing} 
\vspace{-0.75em}

\noindent Importantly, while there appears to be a degree of consensus that these postulates should be strengthened, there is no agreement as to how this should be done. Popular options include the principles respectively associated with the operators of natural revision $\astN$  \cite{boutilier1996iterated}, restrained revision $\astR$ \cite{booth2006admissible} and lexicographic revision $\astL$ \cite{nayak2003dynamic}, semantically defined as follows:

\begin{restatable}{defo}{elemdef}
\label{elemdef}
The operators  $\astN$,  $\astR$ and  $\astL$ are such that:
\begin{itemize}

\item[] $x \preccurlyeq_{\Psi \astN A} y$ iff (1)  $x \in \min(\preccurlyeq_{\Psi}, \mods{A})$, or
(2) $x, y \notin \min(\preccurlyeq_{\Psi}, \mods{A})$ and $x \preccurlyeq_{\Psi} y$

\item[] $x \preccurlyeq_{\Psi \astR A}  y$ iff (1) $x \in \min(\preccurlyeq_{\Psi}, \mods{A})$, or (2)  $x, y \notin \min(\preccurlyeq_{\Psi}, \mods{A})$ and either (a) $x \prec_{\Psi} y$ or (b) $x \sim_{\Psi} y$ and ($x\in\mods{A}$ or $y\in\mods{\neg A}$)

\item[] $x \preccurlyeq_{\Psi \astL A}  y$ iff (1)  $x\in\mods{A}$ and $y\in\mods{\neg A}$, or (2) ($x\in\mods{A}$ iff $y\in\mods{A}$) and $x \preccurlyeq_{\Psi} y$.

\end{itemize}
\end{restatable}
\noindent The suitability of all three operators, which we will group here under the heading of `{\em elementary} revision operators' \cite{DBLP:conf/lori/Chandler019}, has been called into question. Indeed, they assume that a state $\Psi$ can be identified with its corresponding TPO $\preccurlyeq_{\Psi}$ and that belief revision functions map pairs of TPOs and sentences onto TPOs. (For this reason, we will sometimes abuse language and notation and speak, for instance, of the lexicographic revision of a TPO rather than of a state.) But this assumption has been criticised as implausible, with \cite{DBLP:journals/jphil/BoothC17} providing a number of counterexamples.

Accordingly, \cite{DBLP:conf/kr/0001C18} propose a strengthening of the DP postulates that is weak enough to avoid an identification of states with TPOs and is consistent with the characteristic postulates of both $\astR$ and $\astL$ (albeit not of $\astN$). They suggest associating states with structures that are richer than TPOs: `proper ordinal interval (POI) assignments'.

%=====================================================

\section{Conditional revision}
\label{s:CRevision}

%=====================================================

We now turn to our question of interest: How might one extend a revision operator to handle  conditionals as inputs? We shall call such an extended operator, which maps pairs of states and consistent sentences in $L_c$ onto states, a {\em conditional revision} operator. 

In view of the considerable disagreement regarding revision that we noted in the previous section, it would be desirable to find a solution that abstracts from some of the details regarding how this problem is handled. In what follows, we shall propose a method that achieves just this. The  idea that we will exploit is that the result of a conditional revision by a Ramsey Test conditional is determined by that of a plain revision by its corresponding material conditional. More specifically, we will be suggesting the following kind of procedure for constructing $\preccurlyeq_{\Psi\ast A \shortTo B}$:  
\begin{itemize}

\item[] (1) Determine $\preccurlyeq_{\Psi\ast A \shortto B}$.

\item[] (2) Remain as `close' to this TPO as possible, while:
\begin{itemize}

\item[] (a) ensuring that $A\shortTo B\in\cbel{\Psi\ast A\shortTo B}$, and

\item[] (b) retaining some of  $\preccurlyeq_{\Psi\ast A \shortto B}$'s relevant features.
\end{itemize}
\end{itemize}

\noindent Our proposal then is to derive $\preccurlyeq_{\Psi\ast A \shortTo B}$ from $\preccurlyeq_{\Psi\ast A \shortto B}$, via distance minimisation under constraints. Importantly, this suggestion  {\em does not tie us to any particular revision operator}, since it takes $\preccurlyeq_{\Psi\ast A \shortto B}$ as its starting point, irrespective of how it is arrived at.

%=====================================================

\subsection{Distance-minimisation under constraints}
\label{ss:Aproposal}

\noindent  In an early paper on conditional revision,  \cite{DBLP:conf/ecai/NayakPFP96} suggest that the task of conditional revision is no different from that of revision by the corresponding material conditional. Indeed, they note that, on their view of rational revision, whereby they identify $\ast$ with lexicographic revision $\astL$, revision by the material conditional is sufficient to ensure that the corresponding conditional is included in the resulting conditional belief set. In other words, identifying $\ast A\shortTo B$ with $\astL A\shortto B$ is sufficient to secure the following desirable property of `{\em Success}' for conditional revisions:

\begin{tabbing}
\=BLAHHHI \=\kill

\> \SRevR  \>  $\min(\preccurlyeq_{\Psi\ast A\shortTo B}, \mods{A})\subseteq \mods{B}$   \\[0.1cm]

\> {\SRevS}  \>  $A \shortTo B \in \cbel{\Psi\ast A \shortTo B}$   \\[-0.25em]

\end{tabbing} 
\vspace{-0.75em}

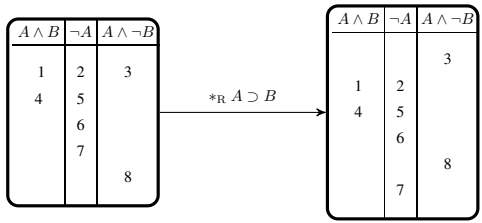
\begin{figure}[h]
\begin{centering}
\begin{tikzpicture}[->,>=stealth']
 \node[state, 
   text width=2cm] (K) 
 {
 \scalebox{0.6}{
\begin{tabular}[c]{c|c|c}
     \multicolumn{1}{c|}{$A\wedge B$} &  $\neg A$   &   \multicolumn{1}{c}{$A\wedge \neg B$} \\
    \hline
    \rule{0pt}{2em}
     \tikzmark{T1}1\tikzmark{B1}   &    \tikzmark{T2}2\tikzmark{B2}  & \tikzmark{T3}3\tikzmark{B3} \\
     \tikzmark{T4}4\tikzmark{B4}    & \tikzmark{T5}5\tikzmark{B5} &    \\
        &    \tikzmark{T6}6\tikzmark{B6}     &  \\
         &    \tikzmark{T7}7\tikzmark{B7}      &    \\
          &       &      \tikzmark{T8}8\tikzmark{B8}      \\[1em]        
  \end{tabular}
}
};

% Next node: astR
 \node[state,       % layout (defined above)
 node distance=4.25cm,     % distance to K
  text width=2cm,
right of=K,        % Position is to the right of K
 yshift=+0cm] (astR)    % move 3cm in y
 {%                     % posistion relative to the center of the 'box'
  \scalebox{0.6}{
\begin{tabular}[c]{c|c|c}
     \multicolumn{1}{c|}{$A\wedge B$} &  $\neg A$   &   \multicolumn{1}{c}{$A\wedge \neg B$} \\
    \hline
    \rule{0pt}{2em}
      &    & \tikzmark{T3}3\tikzmark{B3} \\
     \tikzmark{T1}1\tikzmark{B1}   &    \tikzmark{T2}2\tikzmark{B2}  & \\
     \tikzmark{T4}4\tikzmark{B4}    & \tikzmark{T5}5\tikzmark{B5} &    \\
        &    \tikzmark{T6}6\tikzmark{B6}     &  \\
          &       &      \tikzmark{T8}8\tikzmark{B8}      \\
         &    \tikzmark{T7}7\tikzmark{B7}      &    \\[1em]      
  \end{tabular}
}
};

 % draw the paths and and print some Text below/above the graph
 \path (K) edge  node[anchor=north, above]
                   {
                   \scalebox{0.6}{
                   $\astR~A\shortto B$
                   }
                   } (astR)
;

\end{tikzpicture}

\caption{Illustration of $\min(\preccurlyeq_{\Psi\ast A\shortto B}, \mods{A})\nsubseteq \mods{B}$ with $\ast=\astR$.  The relation $\preceq_{\Psi}$ orders the worlds---depicted by numbers---from bottom to top, with the minimal world on the lowest level. The columns group worlds according to the sentences that they validate. We can see that, here, $\min(\preccurlyeq_{\Psi\astR A\shortto B}, \mods{A})=\{8\}\subset\mods{\neg B}$.}   
\label{fig:NoSuccess}
\end{centering}
\end{figure}

\noindent Since we have, in Section \ref{s:Revision}, rejected identifying rational revision with  lexicographic revision, Nayak {\em et al's} proposal is not on the cards for us. But one might still wonder whether there exists a more acceptable conception of iterated revision that, like lexicographic revision, allows us to meet the requirement of Success by simply revising by the material conditional. But it is easy to find counterexamples to the inclusion $\min(\preccurlyeq_{\Psi\ast A\shortto B}, \mods{A})\subseteq \mods{B}$ for the best known strengthenings of the DP postulates (Figure \ref{fig:NoSuccess} provides a case in point for restrained revision). In fact, we can easily show that, given mild conditions, lexicographic revision is the only operator that fits the bill:  	
%Plausibly, at the level of posterior conditional beliefs, the task of conditional revision differs from that of revision by the corresponding material conditional. In other words, it appears quite reasonable to claim that, generally speaking,  $\preccurlyeq_{\Psi\ast A\shortTo B} \neq \preccurlyeq_{\Psi\ast A\shortto B}$ or equivalently $\cbel{\Psi\ast A\shortTo B} \neq \cbel{\Psi\ast A\shortto B}$.
%% or, equivalently, $\cbel{\Psi\ast A\shortTo B} \neq \cbel{\Psi\ast A\shortto B}$. 
%%  or, equivalently, $A \shortTo B \notin \cbel{\Psi\ast A \shortto B}$. 
%Indeed, we want  `{\em Success}' to hold for conditional revisions:
%
%\begin{tabbing}
%\=BLAHHHI \=\kill
%
%\> \SRevR  \>  $\min(\preccurlyeq_{\Psi\ast A\shortTo B}, \mods{A})\subseteq \mods{B}$   \\[0.1cm]
%
%\> {\SRevS}  \>  $A \shortTo B \in \cbel{\Psi\ast A \shortTo B}$   \\[-0.25em]
%
%\end{tabbing} 
%\vspace{-0.75em}
%\noindent  However, most proposed strengthenings of the DP postulates are such that, generally, $\min(\preccurlyeq_{\Psi\ast A\shortto B}, \mods{A})\nsubseteq \mods{B}$: revision by $A\shortto B$ does not suffice to achieve inclusion of $A\shortTo B$ in the resulting belief set. Figure \ref{fig:NoSuccess} illustrates this fact by reference to restrained revision.

%
%An exception to this is lexicographic revision and indeed, in their treatment of the issue of conditional revision, ??? have suggested to simply identify $\preccurlyeq_{\Psi\ast A\shortTo B}$ with $\preccurlyeq_{\Psi\ast A\shortto B}$.

\begin{restatable}{prop}{OnlyLex}
\label{OnlyLex}
If $\ast$ satisfies AGM, \CRevR{1}, \CRevR{2}, $(\mathrm{Eq}^\ast_\preccurlyeq)$, and the principle according to which, for all $A, B\in L$ and $\Psi\in\mathcal{S}$, $A \shortTo B \in \cbel{\Psi\ast A \shortto B}$, then $\ast =\astL$.
\end{restatable}

\noindent Short of endorsing lexicographic revision, then, revision by the corresponding material conditional is not sufficient for the inclusion of a conditional in the resulting belief set. So just as conditional beliefs can be viewed as  `beliefs in material conditionals plus',  we could say that:
\begin{itemize}
\item[] `Conditional revision is revision by material conditionals plus.' 
\end{itemize} 
% This suggests the following generic semantic procedure for constructing $\preccurlyeq_{\Psi\ast A \shortTo B}$:  
%\begin{itemize}
%
%\item[] (1) Determine $\preccurlyeq_{\Psi\ast A \shortto B}$. 
%
%
%\item[] (2) Remain as `close' to this TPO as possible, while:
%\begin{itemize}
%
%\item[] (a) ensuring that \SRevR~is satisfied, and  
%
%
%\item[] (b) retaining some of  $\preccurlyeq_{\Psi\ast A \shortto B}$'s relevant features.
%\end{itemize}
%\end{itemize}

%\noindent Importantly, note that this suggestion  {\em does not tie us to any particular revision operator}, since it takes $\preccurlyeq_{\Psi\ast A \shortto B}$ as its starting point, irrespective of how it is arrived at. 
%
%To obtain a definite proposal from this, we need a measure of distance between TPOs and a specification of the features of  $\preccurlyeq_{\Psi\ast A \shortto B}$ that are to be retained.

%\footnote{\jake{Note on the use of distance-based solutions in the probabilistic case}}  

%Put semantically, our above proposal amounts to:
%\begin{itemize}
%
%\item[] $\preccurlyeq_{\Psi\ast A\shortTo B}$ is  the TPO that minimizes the distance $d_K$ to $\preccurlyeq_{\Psi\ast A\shortto B}$, given constraints (S$_{\preccurlyeq})$ and  \pprincRevR{1}--\pprincRevR{4} 
%
%\end{itemize}
%
% Can we say more about $\preccurlyeq_{\Psi\ast A\shortTo B}$?

%%=====================================================
%\section{A more specific proposal}
%
%\label{s:RResult}
%
%%=====================================================

\noindent How, then, might we plausibly modify $\preccurlyeq_{\Psi\ast A\shortto B}$ so as to arrive at a TPO $\preccurlyeq_{\Psi\ast A\shortTo B}$ that satisfies \SRevR? 

Satisfaction of this principle, of course, will require some worlds in $\mods{A\wedge B}$ to be promoted in the ranking, notably in relation to certain worlds in $\mods{A\wedge \neg B}$. But we must be cautious as to how this is to take place. Plausibly, for instance, it should not occur at the expense of the worlds in $\mods{\neg A}$. In fact, it seems quite reasonable that, more broadly, the internal ordering of $\mods{A\shortto B}$ should be left untouched. We therefore suggest supplementing  \SRevR~with the following `retainment' principle, which ensures the preservation of these features of $\preccurlyeq_{\Psi\ast A\shortto B}$:

\begin{tabbing}
\=BLAHHHI \=\kill
\> \pprincRevR{1} \> If $x, y \in \mods{A\shortto B}$, then $x \preccurlyeq_{\Psi\ast A\shortTo B}y$ iff \\
\>\> $x \preccurlyeq_{\Psi\ast A\shortto B}y$\\[-0.25em]
\end{tabbing} 
\vspace{-0.75em}

\noindent Its syntactic counterpart is given as follows:

\begin{restatable}{prop}{SyntacticPprincOne}
\label{SyntacticPprincOne}
Given AGM, \pprincRevR{1} is equivalent to 
\begin{tabbing}
\=BLAHHHI \=\kill
\> \pprincRev{1} \> If $A\shortto B\in\Cn(C)$, then $\bel{(\Psi\ast A\shortTo B)\ast C}=$\\
\>\> $\bel{(\Psi\ast A\shortto B)\ast C}$\\[-0.25em]
\end{tabbing} 
\vspace{-0.75em}
\end{restatable}

\noindent Given the DP postulates, this constraint obviously translates into one that connects  $\preccurlyeq_{\Psi} $ and $\preccurlyeq_{\Psi\ast A\shortTo B}$ and whose syntactic counterpart is easily inferable from Proposition \ref{SyntacticPprincOne}:

\begin{restatable}{prop}{PropertiesTwoA}
\label{PropertiesTwoA}
Given \CRevR{1}, \pprincRevR{1} is equivalent to:

\begin{tabbing}
\=BLAHHHI \=\kill
\> \pprincRevR{1'} \> If $x, y \in \mods{A\shortto B}$, then $x \preccurlyeq_{\Psi\ast A\shortTo B}y$ iff \\
\>\> $x \preccurlyeq_{\Psi}  y$\\[-0.25em]
\end{tabbing} 
\vspace{-0.75em}

\end{restatable}

\noindent That conditional revision does not affect the internal ordering of $\mods{A\wedge B}$ or  of $\mods{\neg A}$ is in fact  required by a set of principles for conditional revision proposed in  \cite{DBLP:conf/ijcai/Kern-Isberner99}, to which we shall return later. Our principle adds to these  the constraint that conditional revision by $A\shortTo B$ does not affect the relative standing of worlds in $\mods{\neg A}$ in relation to worlds in $\mods{A\wedge B}$. This further restriction yields the correct verdict in the following scenario:

\begin{example}\label{ex:AkiraOne}
My friend and I have taken our preschoolers Akira and Bashir on holiday. They slept in bunkbeds last night. Since both beds were unmade by the morning, I initially believe that they did not choose to sleep in the same bed but suspend judgment as to which respective beds they did choose. Furthermore, in the event of coming to believe that they in fact did decide to share a bed, I would suspend judgment as to which bed they opted for. I then find out that, if Akira slept on top, then Bashir would have done so too (because he does not like people sleeping above him). What changes? Plausibly, my beliefs will change in the following respect: since I will still believe that they did not share a bed, I will now infer that Akira slept on the bottom bed and Bashir on the top. What of my conditional beliefs? Plausibly, we will have the following continuity: It will remain the case that, were to  find out that they in fact decided to share a bed, I would suspend judgment as to which bed they chose.  
\end{example}

\noindent Indeed, let $A$ and $B$ respectively stand for Akira and for Bashir's sleeping on the top bed and $\Psi$ be my initial state. Assume for simplicity that the set of atomic propositions in $L$ is  $\{A, B\}$. Let $\mods{A\wedge\neg B}=\{x\}$, $\mods{\neg A\wedge B}=\{y\}$, $\mods{A\wedge B}=\{z\}$ and $\mods{\neg A\wedge \neg B}=\{w\}$. We then have $\preccurlyeq_{\Psi}$ plausibly given by $x \sim_{\Psi} y \prec_{\Psi} z \sim_{\Psi} w$. Since $z\sim_{\Psi} w$, our principle entails the plausible result that $z\sim_{\Psi \ast A\shortTo B} w$.

But unfortunately, \SRevR~and \pprincRevR{1} are not generally jointly sufficient to have the TPO $\preccurlyeq_{\Psi\ast A\shortto B}$ determine the TPO $\preccurlyeq_{\Psi\ast A\shortTo B}$. Our suggestion is to  close the gap by means of distance minimisation. More specifically, we propose to consider the closest TPO that satisfies--or, in the event of a tie, some aggregation of the closest TPOs that satisfy--our two constraints. 

In terms of measuring the distance between TPOs, a natural choice is the so-called Kemeny distance $d_K$:

\begin{restatable}{defo}{SymDiff}
\label{SymDiff}
 $d_K(\preccurlyeq, \preccurlyeq'):= \lvert (\preccurlyeq - \preccurlyeq') \cup (\preccurlyeq' - \preccurlyeq) \rvert$.
\end{restatable}

\noindent Informally, $d_K(\preccurlyeq, \preccurlyeq')$ returns the number of disagreements over relations of weak preference between the two orderings, returning the number of pairs that are in $\preccurlyeq$ but not in $\preccurlyeq'$ and vice versa. This  measure is standard fare in the social choice literature. It was introduced there in \cite{10.2307/20026529} and received an axiomatisation in terms of a set of prima facie attractive properties in \cite{KemenySnell62}. 

In the section that follows we shall show that there exists a unique $d_K$-closest TPO that meets the requirements \SRevR~and \pprincRevR{1}, which can be obtained from $\preccurlyeq_{\Psi\ast A \shortto B}$ in a simple and familiar manner.

%=====================================================
\subsection{A construction of the posterior TPO}

\label{ss:RResult}

%=====================================================

To outline our main result, we first need the following item of notation (see Figure \ref{fig:Downset} for illustration):

\begin{restatable}{defo}{Downset} 
\label{Downset}
For any sentence $A\in L$ and TPO $\preccurlyeq$, we denote by $D(\preccurlyeq, A)$ the {\em down-set} of the members of $\min(\preccurlyeq, \mods{A})$. It is given by $D(\preccurlyeq, A):=\{x\mid x\preccurlyeq z \mathrm{,~for~some~} z\in\min(\preccurlyeq, \mods{A})\}$. 
\end{restatable}

\begin{figure}[!h]
\begin{centering}
\begin{tikzpicture}[->,>=stealth']
\scalebox{1}{%
 \node[state,
  text width=2cm,        % max text width
 ] (astN)   
 {
 \scalebox{0.6}{
\begin{tabular}[c]{c|c|c}
     \multicolumn{1}{c|}{$A\wedge B$} &  $\neg A$   &   \multicolumn{1}{c}{$A\wedge \neg B$} \\
    \hline
    \rule{0pt}{2em}
      &    & \tikzmark{T3}3\tikzmark{B3} \\
     \tikzmark{T1}1\tikzmark{B1}   &    \tikzmark{T2}2\tikzmark{B2}  & \\
     \tikzmark{T4}4\tikzmark{B4}    & \tikzmark{T5}5\tikzmark{B5} &    \\
        &    \tikzmark{T6}6\tikzmark{B6}     &  \\
          &       &      \tikzmark{T8}8\tikzmark{B8}      \\
         &    \tikzmark{T7}7\tikzmark{B7}      & \tikzmark{T10}~\tikzmark{B10}    \\[1em]  
  \end{tabular}
\DrawBox[thick, black, dashed]{T4}{B4}
\DrawBox[thick, black]{T4}{B10}
}
};

}
\end{tikzpicture}

\caption{Down-set $D(\preccurlyeq_{\Psi\ast A\shortto B}, A\wedge B)$ of the members of $\min(\preccurlyeq_{\Psi\ast A\shortto B}, \mods{A\wedge B})$. The set $\min(\preccurlyeq_{\Psi\ast A\shortto B}, \mods{A\wedge B})$, which here is a singleton,  is marked by a dashed box. $D(\preccurlyeq_{\Psi\ast A\shortto B}, A\wedge B)$ is marked by a solid box.}   
\label{fig:Downset}
\end{centering}
\end{figure}

\noindent With this in hand, we propose:  

\begin{restatable}{defo}{MyProposal} 
\label{MyProposal}
Let $\ast$ be a function from $\mathbb{S}\times L$ to $\mathbb{S}$. Then we denote by $\circledast$ an arbitrary extension of $\ast$ to the domain $\mathbb{S}\times L_c$, such that $\preccurlyeq_{\Psi \circledast A\shortTo B}$ is given by the lexicographic revision of $\preccurlyeq_{\Psi \ast A\shortto B}$ by $D(\preccurlyeq_{\Psi\ast A\shortto B}, A\wedge B)\cap\mods{A\shortto B}$.\footnote{In case we identify states with TPOs, there will exist only one such extension.}
%
%\begin{itemize}
%
%\item[] $x \preccurlyeq_{\Psi \ast A\shortTo B}  y$ iff $x \preccurlyeq_{(\Psi \ast A\shortto B)\astL D(\preccurlyeq_{\Psi\ast A\shortto B}, A\wedge B)\cap\mods{A\shortto B}}  y$
%
%\end{itemize}
\end{restatable}

\noindent  The operator so-defined is illustrated in Figure \ref{fig:RevCon}, which depicts the resulting relation between $\preccurlyeq_{\Psi\ast A\shortto B}$ and $\preccurlyeq_{\Psi\circledast A\shortTo B}$. Interestingly, in the special case of a Ramsey Test conditional with a tautologous antecedent, this transformation of $\preccurlyeq_{\Psi\ast A\shortto B}$ amounts to its natural revision by the consequent.

\begin{figure}[!h]
\begin{centering}
\begin{tikzpicture}[->,>=stealth']
\scalebox{1}{%
 \node[state,
  text width=2cm,        % max text width
 ] (astRandom)   
 {
 \scalebox{0.6}{
\begin{tabular}[c]{c|c|c}
     \multicolumn{1}{c|}{$A\wedge B$} &  $\neg A$   &   \multicolumn{1}{c}{$A\wedge \neg B$} \\
    \hline
    \rule{0pt}{2em}
      &    & \tikzmark{T3}3\tikzmark{B3} \\
     \tikzmark{T1}1\tikzmark{B1}   &    \tikzmark{T2}2\tikzmark{B2}  & \\
     \tikzmark{T4}4\tikzmark{B4}    & \tikzmark{T5}5\tikzmark{B5} &    \\
        &    \tikzmark{T6}6\tikzmark{B6}     &  \\
          &       &      \tikzmark{T8}8\tikzmark{B8}      \\
         &    \tikzmark{T7}7\tikzmark{B7}      &    \\[1em]  
  \end{tabular}
\DrawBox[thick, centuryred]{T4}{B7}
%\DrawBox[thick, centuryred]{T6}{B9}
}
};

 \node[state,     % layout (defined above)
 node distance=4.25cm,     % distance to K
 text width=2cm,        % max text width
right of=astRandom,        % Position is to the right of K
 yshift=+0cm] (astRandomMod)   
 {
 \scalebox{0.6}{
\begin{tabular}[c]{c|c|c}
     \multicolumn{1}{c|}{$A\wedge B$} &  $\neg A$   &   \multicolumn{1}{c}{$A\wedge \neg B$} \\
    \hline
    \rule{0pt}{2em}
      &    & \tikzmark{T3}3\tikzmark{B3} \\
     \tikzmark{T1}1\tikzmark{B1}   &    \tikzmark{T2}2\tikzmark{B2}  & \\
          &       &      \tikzmark{T8}8\tikzmark{B8}      \\
     \tikzmark{T4}4\tikzmark{B4}    & \tikzmark{T5}5\tikzmark{B5} &    \\
        &    \tikzmark{T6}6\tikzmark{B6}     &  \\
         &    \tikzmark{T7}7\tikzmark{B7}      &    \\[1em]   
  \end{tabular}
\DrawBox[thick, centuryred]{T4}{B7}
%\DrawBox[thick, centuryred]{T6}{B9}
}
};

 \path (astRandom) edge[dashed]  node[anchor=west, right]
                   {
                   ~
                   } (astRandomMod)
;

}
\end{tikzpicture}

\caption{Relation between $\preccurlyeq_{\Psi\ast A\shortto B}$ (depicted on the left) and $\preccurlyeq_{\Psi\circledast A\shortTo B}$ (depicted on the right).  $D(\preccurlyeq_{\Psi\ast A\shortto B}, A\wedge B)\cap\mods{A\shortto B}$ is marked by a box.}   
\label{fig:RevCon}
\end{centering}
\end{figure}

\noindent We propose to identify $\preccurlyeq_{\Psi\ast A\shortTo B}$ with $\preccurlyeq_{\Psi\circledast A\shortTo B}$. We do so on the basis of our main technical result, which is:

\begin{restatable}{thm}{Representation}
\label{Representation}
The unique TPO that minimises the distance $d_K$ to $\preccurlyeq_{\Psi\ast A\shortto B}$, given constraints \SRevR~and \pprincRevR{1} is given by $\preccurlyeq_{\Psi\circledast A\shortTo B}$.
\end{restatable}

\noindent As indicated above, short of endorsing lexicographic revision, which we do not want to do, the constraint of Success prevents us from having $\cbel{\Psi\ast A\shortTo B} = \cbel{\Psi\ast A\shortto B}$ for all $A, B\in L$, $\Psi\in\mathcal{S}$. Having said that, a restricted version of this equality does hold for our proposal  in the form of the following plausible `Vacuity' postulate, which tells us that if revision by the material conditional leads to the conditional being accepted, then it is revision enough:

\begin{tabbing}
\=BLAHHHI \=\kill
\> $(\mathrm{V}^\ast)$ \> If $A\shortTo B\in\cbel{\Psi\ast A\shortto B}$, then\\
\>\> $\cbel{\Psi\ast A\shortTo B}=\cbel{\Psi\ast A\shortto B}$\\[-0.25em]
\end{tabbing} 
\vspace{-0.75em}

\noindent Furthermore, as a consequence of one of the results established in the proof of Theorem \ref{Representation}, we can also derive an interesting minimal change result with a more syntactic flavour:

\begin{restatable}{prop}{SyntacticMinimality}
\label{SyntacticMinimality}
Let $\ast$ be a function from $\mathbb{S}\times L$ to $\mathbb{S}$ and $\ast'$ an extension of $\ast$ to the domain $\mathbb{S}\times L_c$, satisfying \SRevR~and  \pprincRevR{1}. Then, if  $\cbel{\Psi\ast' A\shortTo B}$  agrees with $\cbel{\Psi\ast A\shortto B}$ on all conditionals with a given antecedent $C$, so does $\cbel{\Psi\circledast A\shortTo B}$

\end{restatable}

%=====================================================
\subsection{Some general features}

\label{ss:AltChar}

%=====================================================

\noindent We have seen that our proposal to handle conditional revision using distance minimisation under constraints yields a unique TPO that can be obtained via lexicographic revision of $\preccurlyeq_{\Psi\ast A\shortto B}$ by a particular proposition. In this section, we discuss some of its general consequences, including three additional retainment principles that it implies.

It is easy to establish the following:

\begin{restatable}{prop}{RetSoundness}
\label{RetSoundness}
Let $\ast$ be a function from $\mathbb{S}\times L_c$ to $\mathbb{S}$. Then, if $\ast=\circledast$, then  $\ast$ satisfies: 
\begin{tabbing}
\=BLAHHHI \=\kill
\> \pprincRevR{2} \> If $x, y \in \mods{A\wedge \neg B}$,  then $x \preccurlyeq_{\Psi\ast A\shortTo B}y$ iff\\
\>\> $x \preccurlyeq_{\Psi\ast A\shortto B}y$ \\[0.1cm]
\> \pprincRevR{3} \> If $x\in \mods{A\shortto B}$, $y\in \mods{A\wedge \neg B}$,  and  \\
\>\> $x \prec_{\Psi\ast A\shortto B}y$, then $x \prec_{\Psi\ast A\shortTo B}y$  \\[0.1cm]
\> \pprincRevR{4} \> If $x\in \mods{A\shortto B}$, $y\in \mods{A\wedge \neg B}$,  and \\
\>\> $x \preccurlyeq_{\Psi\ast A\shortto B}y$, then $x \preccurlyeq_{\Psi\ast A\shortTo B}y$\footnotemark{} \\[-0.25em]
\end{tabbing} 
\vspace{-0.75em}
\end{restatable}

\footnotetext{It turns out that this set of principles is sufficient to fully characterise our proposal when supplemented with \SRevR, \pprincRevR{1} and the two following further postulates:
\begin{tabbing}
\=BLAHHHI \=\kill
\> \pprincRevR{5} \> If $x\in \mods{A\wedge \neg B}$,  $y\notin D(\preccurlyeq_{\Psi\ast A\shortto B}, A\wedge B)$,\\
\>\> $y\in \mods{A\shortto B}$, and $x \prec_{\Psi\ast A\shortto B}y$, then $x \prec_{\Psi\ast A\shortTo B}y$  \\[0.1cm]
\> \pprincRevR{6} \> If $x\in \mods{A\wedge \neg B}$, $y\notin D(\preccurlyeq_{\Psi\ast A\shortto B}, A\wedge B)$,\\
\>\> $y\in \mods{A\shortto B}$, and $x \preccurlyeq_{\Psi\ast A\shortto B}y$, then $x \preccurlyeq_{\Psi\ast A\shortTo B}y$  \\[-0.25em]
\end{tabbing} 
\vspace{-0.75em}
These, however, are not immediately as clearly interpretable. 
}

\noindent The conjunction of \pprincRevR{1}, with these three principles simply  tells us that the only admissible transformations, when  moving from $\preccurlyeq_{\Psi\ast A\shortto B}$ to $\preccurlyeq_{\Psi\ast A\shortTo B}$, involve a doxastic  `demotion' of worlds in  $\mods{A\wedge \neg B}$ in relation to worlds in $\mods{A\shortto B}$, raising, in the ordering, the position of the former in relation to the latter. They have a similar flavour to that of the DP postulates, which tell us that that the only admissible transformations, when  moving from $\preccurlyeq_{\Psi}$ to $\preccurlyeq_{\Psi\ast A}$, involve a demotion of worlds in  $\mods{\neg A}$ in relation to worlds in $\mods{A}$.

We note the immediate implications of these principles, in the presence of the DP postulates:

\begin{restatable}{prop}{PropertiesTwo}
\label{PropertiesTwo}
Given \CRevR{1}--\CRevR{4}, \pprincRevR{2} holds iff:

\begin{tabbing}
\=BLAHHHI \=\kill
\> \pprincRevR{2'} \> If $x, y \in \mods{A\wedge \neg B}$,  $x \preccurlyeq_{\Psi\ast A\shortTo B}y$ iff $x \preccurlyeq_{\Psi}  y$ \\[-0.25em]
\end{tabbing} 
\vspace{-0.75em}

\noindent and \pprincRevR{3} and  \pprincRevR{4} respectively entail:

\begin{tabbing}
\=BLAHHHI \=\kill
\> \pprincRevR{3'} \> If $x\in \mods{A\shortto B}$, $y\in \mods{A\wedge \neg B}$, and $x \prec_{\Psi} y$, \\
\>\>then $x \prec_{\Psi\ast A\shortTo B}y$  \\[0.1cm]
\> \pprincRevR{4'} \> If $x\in \mods{A\shortto B}$, $y\in \mods{A\wedge \neg B}$, and $x \preccurlyeq_{\Psi}  y$,  \\
\>\> then $x \preccurlyeq_{\Psi\ast A\shortTo B}y$  \\[-0.25em]
\end{tabbing} 
\vspace{-0.75em}

\noindent but the converse entailments do not hold. 

\end{restatable}

\noindent The syntactic counterparts of \pprincRevR{2}--\pprincRevR{4} are given in the following proposition, with the counterparts of \pprincRevR{2'}--\pprincRevR{4'} being easily inferable from these:
\begin{restatable}{prop}{SyntacticPprinc}
\label{SyntacticPprinc}
Given AGM, \pprincRevR{2}--\pprincRevR{4} are respectively equivalent to 
\begin{tabbing}
\=BLAHHHI \=\kill
\> \pprincRev{2} \> If $A\wedge \neg B\in\Cn(C)$, then  $\bel{(\Psi\ast A\shortTo B)\ast C}=$\\
\>\> $\bel{(\Psi\ast A\shortto B)\ast C}$\\[0.1cm]
\> \pprincRev{3} \> If $A\shortto B\in\bel{(\Psi\ast A\shortto B)\ast C}$, then $A\shortto B\in$\\
\>\> $\bel{(\Psi\ast A\shortTo B)\ast C}$\\[0.1cm]
\> \pprincRev{4} \> If $A\wedge\neg  B\notin\bel{(\Psi\ast A\shortto B)\ast C}$, then $A\wedge\neg  B\notin$\\
\>\> $\bel{(\Psi\ast A\shortTo B)\ast C}$\\[-0.25em]
\end{tabbing} 
\vspace{-0.75em}
\end{restatable}

\noindent 
%
%Taken in conjunction with \pprincRev{1}, \pprincRev{2} says that any difference between revising by a conditional and revising by its material counterpart can only show up, upon further revision, when the input to that further revision is consistent with both the truth and the falsity of the relevant material conditional.  \jake{
%\pprincRev{3} and \pprincRev{4}  tell us that revising by a conditional rather than by its material counterpart doesn't worsen the doxastic status that this material conditional would have upon any further revision.
%} 
%
In introducing \pprincRev{1} above, we noted that, given \CRevR{1}, it strengthens, in a plausible manner, part of a principle proposed in \cite{DBLP:conf/ijcai/Kern-Isberner99}. 
 It turns out that, in the presence of the full set of DP postulates, \pprincRev{1}--\pprincRev{4}  enable us to recover the trio of principles proposed by Kern-Isberner. These ``KI postulates'',  originally named ``(CR5)'' to``(CR7)'', are given semantically by: 
\begin{tabbing}
\=BLAHHHI \=\kill
\> {(KI1$_{\preccurlyeq}^{\ast}$)} \> If $x,y\in \mods{A\wedge B}$, $x,y\in \mods{\neg A}$ \\
\>\> or  $x,y\in \mods{A\wedge \neg B}$, then $x \preccurlyeq_{\Psi} y$ iff \\
\>\> $x \preccurlyeq_{\Psi\ast A\shortTo B}y$ \\[0.1cm]
\>  {(KI2$_{\preccurlyeq}^{\ast}$)} \> If $x\in \mods{A\wedge B}$, $y\in \mods{A\wedge \neg B}$ and $x \prec_{\Psi} y$,\\
\>\> then $x \prec_{\Psi\ast A\shortTo B}y$ \\[0.1cm]
\>  {(KI3$_{\preccurlyeq}^{\ast}$)} \> If $x\in \mods{A\wedge B}$, $y\in \mods{A\wedge \neg B}$ and $x \preccurlyeq_{\Psi} y$,\\
\>\> then $x \preccurlyeq_{\Psi\ast A\shortTo B}y$ \\[-0.25em]
\end{tabbing} 
\vspace{-0.75em}

\noindent We can see that \KIRevR{1} follows from \pprincRevR{1} and \pprincRevR{2}, given \CRevR{1} and \CRevR{2}.  \KIRevR{2} follows from the conjunction of \pprincRevR{3} and \CRevR{3}, while \KIRevR{3} follows from the conjunction of \pprincRevR{4} and \CRevR{4}.\footnote{We note that the KI postulates, in turn, subsume the DP postulates, which correspond to the special cases in which $A=\top$. Indeed, \KIRevR{1} yields the conjunction of \CRevR{1} and \CRevR{2}, while \KIRevR{2} and \KIRevR{3} give us \CRevR{3} and \CRevR{4}, respectively.}

In view of Theorem \ref{Representation} and  Proposition \ref{RetSoundness}, it follows that, if a revision operator $\ast$ satisfies \CRevR{1} to \CRevR{4}, then the conditional revision $\circledast$ operator that extends it in the manner described in Definition \ref{MyProposal} satisfies \KIRevR{1} to \KIRevR{3}.

\pprincRevR{3} and \pprincRevR{4} tell us that conditional revision by $A\shortTo B$ preserves any  `good news' for worlds in $\mods{A\shortto B}$, compared to worlds in $\mods{A\wedge \neg B}$, that revision by $A\shortto B$ would bring. Given \CRevR{3}~and \CRevR{4}, they notably add to \KIRevR{2} and \KIRevR{3} the idea that worlds in $\mods{\neg A}$ should not be demoted with respect to worlds in $\mods{A\wedge\neg B}$ in moving from $\preccurlyeq_{\Psi}$ to  $\preccurlyeq_{\Psi \ast A\shortTo B}$. The appeal of this constraint is highlighted in the following case:

\begin{example}
I am due to visit my hometown and would like to catch up with my friends Alex and Ben. Unfortunately, both of them moved away years ago and I doubt that I will see either. If I were to find out of either of them that he was going to be around, I would still believe that the other was not. Furthermore if I were to find out that exactly one of them would be back, I would not be able to guess which one of the two that would be. A friend now tells me that if Alex will be in town, then so will Ben. Very clearly, it should not be the case that, as a result of this new information, I would now take Alex to be a more plausible candidate for being the only one of my two friends that I will see (quite the contrary).
\end{example}

\noindent  Let $A$ and $B$ respectively stand for Alex and for Ben’s being back in town and $\Psi$ be my initial state. Assume for simplicity that the set of atomic propositions in $L$ is simply $\{A,B\}$.  Let $\mods{A\wedge\neg B}=\{x\}$, $\mods{\neg A\wedge B}=\{y\}$, $\mods{A\wedge B}=\{z\}$ and $\mods{\neg A\wedge \neg B}=\{w\}$. We then have $\preccurlyeq_{\Psi}$ plausibly given by $w \prec_{\Psi} x \sim_{\Psi} y \prec_{\Psi} z$. Since $ y \preccurlyeq_{\Psi} x$, our principle entails that $ y \preccurlyeq_{\Psi\ast A\shortTo B} x$, as it intuitively should be.

Aside from entailing the three further retainment principles that we have discussed, we also note that our postulates have the happy consequence of securing the following `Doxastic Equivalence' principle, according to which conditional revisions are indistinguishable from revisions by material conditionals at the level of belief sets:

\begin{tabbing}
\=BLAHHHI \=\kill

\>  \DERevR \>  $\min(\preccurlyeq_{\Psi\ast A\shortTo B}, W) = \min(\preccurlyeq_{\Psi\ast A\shortto B}, W)$ \\[-0.25em]

%\>  \jake{(???)} \>  $\bel{\Psi\ast A\shortTo B} = \bel{\Psi\ast A\shortto B}$ \\[-0.25em]

\end{tabbing} 
\vspace{-0.75em}
More precisely, it is easy to show that:

\begin{restatable}{prop}{NoAddedFactualBeliefs}
\label{NoAddedFactualBeliefs}
\SRevR,  \pprincRevR{1}, \pprincRevR{3} and \pprincRevR{4} collectively entail \DERevR.
\end{restatable}

\subsection{Elementary conditional revision operators}

\label{ss:Elementary}

%=====================================================

A few interesting observations can be made regarding the more specific case in which $\ast$ is an elementary operator (i.e.~belongs to the set $\{\astN, \astR, \astL\}$), which we illustrate in Figure \ref{fig:Elementary}. Having said that, we have noted above our significant reservations about identifying rational revision with any of these operators. This section is therefore addressed to those who are rather more optimistic.

\begin{centering}
\begin{figure}[h]
\begin{tikzpicture}[->,>=stealth']
\scalebox{1}{%
 \node[state,
 text width=2cm,        % max text width
 ] (K) 
 {
 \scalebox{0.6}{
\begin{tabular}[c]{c|c|c}
     \multicolumn{1}{c|}{$A\wedge B$} &  $\neg A$   &   \multicolumn{1}{c}{$A\wedge \neg B$} \\
    \hline
    \rule{0pt}{2em}
     \tikzmark{T1}1\tikzmark{B1}   &    \tikzmark{T2}2\tikzmark{B2}  & \tikzmark{T3}3\tikzmark{B3} \\
     \tikzmark{T4}4\tikzmark{B4}    & \tikzmark{T5}5\tikzmark{B5} &    \\
        &    \tikzmark{T6}6\tikzmark{B6}     &  \\
         &    \tikzmark{T7}7\tikzmark{B7}      &    \\
          &       &      \tikzmark{T8}8\tikzmark{B8}      \\[1em]        
  \end{tabular}
}
};

% Next node: astR
 \node[state,       % layout (defined above)
 node distance=3.5cm,     % distance to K
 text width=2cm,        % max text width
 below of=K,        % Position is to the right of K
 yshift=+0cm] (astR)    % move 3cm in y
 {%                     % posistion relative to the center of the 'box'
  \scalebox{0.6}{
\begin{tabular}[c]{c|c|c}
     \multicolumn{1}{c|}{$A\wedge B$} &  $\neg A$   &   \multicolumn{1}{c}{$A\wedge \neg B$} \\
    \hline
    \rule{0pt}{2em}
      &    & \tikzmark{T3}3\tikzmark{B3} \\
     \tikzmark{T1}1\tikzmark{B1}   &    \tikzmark{T2}2\tikzmark{B2}  & \\
     \tikzmark{T4}4\tikzmark{B4}    & \tikzmark{T5}5\tikzmark{B5} &    \\
        &    \tikzmark{T6}6\tikzmark{B6}     &  \\
          &       &      \tikzmark{T8}8\tikzmark{B8}      \\
         &    \tikzmark{T7}7\tikzmark{B7}      &    \\[1em]       
  \end{tabular}
\DrawBox[thick, centuryred]{T4}{B7}
%\DrawBox[thick, centuryred]{T8}{B9}
}
};

% Next node: asLL
 \node[state,       % layout (defined above)
 node distance=2.5cm,     % distance to K
 text width=2cm,        % max text width
 left of=astR,        % Position is to the right of K
 yshift=+0cm] (astL)    % move 3cm in y
 {%                     % posistion relative to the center of the 'box'
  \scalebox{0.6}{
\begin{tabular}[c]{c|c|c}
     \multicolumn{1}{c|}{$A\wedge B$} &  $\neg A$   &   \multicolumn{1}{c}{$A\wedge \neg B$} \\
    \hline
    \rule{0pt}{2em}
      &   & \tikzmark{T3}3\tikzmark{B3} \\
          &       &      \tikzmark{T8}8\tikzmark{B8}      \\
     \tikzmark{T1}1\tikzmark{B1}   &    \tikzmark{T2}2\tikzmark{B2}  &  \\
     \tikzmark{T4}4\tikzmark{B4}    & \tikzmark{T5}5\tikzmark{B5} &    \\
        &    \tikzmark{T6}6\tikzmark{B6}     &  \\
         &    \tikzmark{T7}7\tikzmark{B7}      &    \\ [1em]     
  \end{tabular}
\DrawBox[thick, centuryred]{T4}{B7}
}
};

% Next node: astN
 \node[state,       % layout (defined above)
 node distance=2.5cm,     % distance to K
 text width=2cm,        % max text width
right of=astR,        % Position is to the right of K
 yshift=+0cm] (astN)    % move 3cm in y
 {
 \scalebox{0.6}{
\begin{tabular}[c]{c|c|c}
     \multicolumn{1}{c|}{$A\wedge B$} &  $\neg A$   &   \multicolumn{1}{c}{$A\wedge \neg B$} \\
    \hline
    \rule{0pt}{2em}
     \tikzmark{T1}1\tikzmark{B1}   &    \tikzmark{T2}2\tikzmark{B2}  & \tikzmark{T3}3\tikzmark{B3} \\
     \tikzmark{T4}4\tikzmark{B4}    & \tikzmark{T5}5\tikzmark{B5} &    \\
        &    \tikzmark{T6}6\tikzmark{B6}     &  \\
          &       &      \tikzmark{T8}8\tikzmark{B8}      \\
             &    \tikzmark{T7}7\tikzmark{B7}      &    \\[1em]   
  \end{tabular}
\DrawBox[thick, centuryred]{T4}{B7}
%\DrawBox[thick, centuryred]{T6}{B9}
}
};

% Next node: astNMod
 \node[state,       % layout (defined above)
 node distance=3.5cm,     % distance to K
 text width=2cm,        % max text width
below of=astN,        % Position is to the right of K
 yshift=+0cm] (astNMod)    % move 3cm in y
 {%                     % posistion relative to the center of the 'box'
  \scalebox{0.6}{
\begin{tabular}[c]{c|c|c}
     \multicolumn{1}{c|}{$A\wedge B$} &  $\neg A$   &   \multicolumn{1}{c}{$A\wedge \neg B$} \\
    \hline
    \rule{0pt}{2em}
     \tikzmark{T1}1\tikzmark{B1}   &    \tikzmark{T2}2\tikzmark{B2}  & \tikzmark{T3}3\tikzmark{B3} \\
          &       &      \tikzmark{T8}8\tikzmark{B8}      \\
     \tikzmark{T4}4\tikzmark{B4}    & \tikzmark{T5}5\tikzmark{B5} &    \\
        &    \tikzmark{T6}6\tikzmark{B6}     &  \\
             &    \tikzmark{T7}7\tikzmark{B7}      &    \\[1em]   
  \end{tabular}
%\DrawBox[thick, red]{T32}{B42}
\DrawBox[thick, centuryred]{T4}{B7}
%\DrawBox[thick, centuryred]{T6}{B9}
}
};

% Next node: astRMod
 \node[state,       % layout (defined above)
 node distance=7cm,     % distance to K
 text width=2cm,        % max text width
 below of=K,        % Position is to the right of K
 yshift=+0cm] (astRMod)    % move 3cm in y
 {%                     % posistion relative to the center of the 'box'
  \scalebox{0.6}{
\begin{tabular}[c]{c|c|c}
     \multicolumn{1}{c|}{$A\wedge B$} &  $\neg A$   &   \multicolumn{1}{c}{$A\wedge \neg B$} \\
    \hline
    \rule{0pt}{2em}
      &    & \tikzmark{T3}3\tikzmark{B3} \\
     \tikzmark{T1}1\tikzmark{B1}   &    \tikzmark{T2}2\tikzmark{B2}  & \\
          &       &      \tikzmark{T8}8\tikzmark{B8}      \\
     \tikzmark{T4}4\tikzmark{B4}    & \tikzmark{T5}5\tikzmark{B5} &    \\
        &    \tikzmark{T6}6\tikzmark{B6}     &  \\
         &    \tikzmark{T7}7\tikzmark{B7}      &    \\[1em]    
  \end{tabular}
\DrawBox[thick, centuryred]{T4}{B7}
%\DrawBox[thick, centuryred]{T8}{B9}
}
};

 % draw the paths and and print some Text below/above the graph
 \path (K) edge  node[anchor=west]
                   {
                   \scalebox{0.6}{
                   $\astR~A\shortto B$
                   }
                   } (astR)
;

 \path (K) edge  node[anchor=west, right]
                   {
                   \scalebox{0.6}{
                   $\astL~A\shortto B$
                   }
                   } (astL.north)
;

 \path (K) edge  node[anchor=west, right]
                   {
                   \scalebox{0.6}{
                   $\astN~A\shortto B$
                   }
                   } (astN.north)

;
 \path (astN) edge[dashed]  node[anchor=west, right]
                   {
                   ~
                   } (astNMod)
;

 \path (astR) edge[dashed]  node[anchor=west, right]
                   {
                   ~
                   } (astRMod)
;

 \path (astL)  	edge[loop below, dashed]    node[anchor=south]{~} (astL)

;

}
\end{tikzpicture}

\vspace{3em}

\caption{Two-step procedure for revision by $A\shortTo B$ according to the respective proposed extensions of the elementary operators. $D(\preccurlyeq_{\Psi\ast A\shortto B}, A\wedge B)\cap\mods{A\shortto B}$ is marked by a box. The first step is denoted by a full arrow and the second by a dashed arrow.}   
\label{fig:Elementary}
\end{figure}
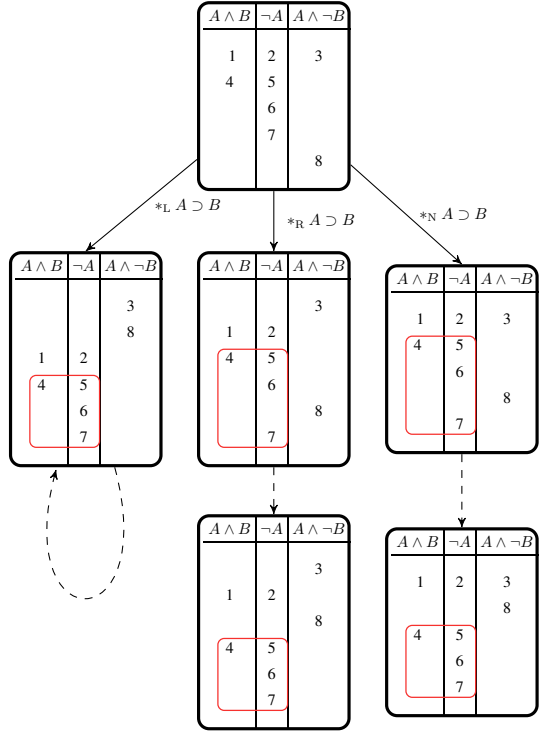
\end{centering}

First, we note that, in two of the special cases of interest, one of our two steps becomes superfluous.

If $\ast=\astL$, then the second step of our procedure is redundant. Indeed, for any $x$ such that $x\in\mods{A\wedge B}$ and any $y\in\mods{A\wedge \neg B}$, we have $x\prec_{\Psi\astL A\shortto B}y$. Hence every world that is in $D(\preccurlyeq_{\Psi\ast A\shortto B}, A\wedge B)\cap\mods{A\shortto B}$ is already strictly more minimal, in $\preccurlyeq_{\Psi\ast A\shortto B}$,  than any world that is not. 

If $\ast=\astN$, then the first step of our procedure plays no role: we would obtain the same result by simply directly applying the second transformation to the initial TPO. This is apparent from the 
fact that natural revision by $A\shortto B$  leaves unaffected the respective internal orderings of $\overline{D(\preccurlyeq_{\Psi\ast A\shortto B}, A\wedge B)}$, $D(\preccurlyeq_{\Psi\ast A\shortto B}, A\wedge B)\cap\mods{A\shortto B}$ and $D(\preccurlyeq_{\Psi\ast A\shortto B}, A\wedge B)\cap\mods{A\wedge \neg B}$, while the latter, on our proposal, jointly determine $\preccurlyeq_{\Psi\ast A \shortTo B}$.

Secondly, in the case of elementary operators more generally, it can be shown that, on our proposal for $\ast A\shortTo B$, the posterior internal ordering $\preccurlyeq_{\Psi\ast A\shortTo B}\cap\mods{A}$ of the set of $A$-worlds is recovered by revising by $B$ the restriction $\preccurlyeq\cap\mods{A}$ of the prior ordering to the $A$-worlds:\footnote{If, that is, we extend in the obvious manner the domain of $\ast$ to cover any TPO over some {\em subset} of $W$.}

\begin{restatable}{prop}{PropertiesThree}
\label{PropertiesThree}
If $\ast$ is an elementary revision operator, then $\ast$ and $\circledast$ satisfy: $\preccurlyeq_{\Psi\circledast A\shortTo B}\cap\mods{A}=(\preccurlyeq_{\Psi}\cap\mods{A})\ast B$.
%\footnote{\jake{change notation for worlds here} The right hand side of this equality is not itself generally equal to $\preccurlyeq_{\Psi\ast B}\cap\mods{A}$. To illustrate, let $\preccurlyeq_{\Psi}$ be given by $\overline{A}\overline{B}\prec_{\Psi} \overline{A}B\prec_{\Psi} A\overline{B}\prec _{\Psi}AB$. Let $\prec'$ denote $(\preccurlyeq_{\Psi}\cap\mods{A})\astN B$. Then $(\preccurlyeq_{\Psi}\cap\mods{A})\astN B$ is given by $AB\prec'A\overline{B}$. But   $\preccurlyeq_{\Psi\astN B}\cap\mods{A}$ is given by $A\overline{B}\prec_{\Psi \astN B} AB$.}

\end{restatable}

\noindent  In other words: if one disregards the worlds in which the antecedent is false, the proposed transformation amounts to revision by the consequent. 
%This strikes us as being a fairly attractive property.

Finally, in \cite[Theorem 4]{DBLP:conf/lori/Chandler019}, it was noted that there is an interesting connection between natural revision and the rational closure operator  $\CRat$   \cite[Defs 20 and 21]{lehmann1992does}, which minimally extends any consistent set of conditionals to a set of conditionals corresponding to a rational consequence relation. This connection was that, if $\neg A\notin\cbel{\Psi}$, then $\cbel{\Psi\astN A}=\CRat(\cbel{\Psi}\cup\{A\})$. This connection deepens on the proposed extension of natural revision to the conditional case. The proof of Chandler \& Booth's theorem can be built upon to establish the following non-trivial result: 

\begin{restatable}{prop}{NatAndRatClosure}
\label{NatAndRatClosure}
If $\ast=\astN$, then, if $A\shortTo \neg B\notin\cbel{\Psi}$, then $\cbel{\Psi\circledast A\shortTo B}=\CRat(\cbel{\Psi}\cup\{A\shortTo B\})$ 
\end{restatable}

%=====================================================

\section{Related research}
\label{s:RResearch}

%=====================================================

We have already presented Kern-Isberner's trio of postulates for conditional revision and briefly discussed (and rejected)  Nayak {\em et al}'s suggestion to treat conditional revision as lexicographic revision by a material conditional. In this section we turn to two further proposals that have been made in the literature and  briefly compare them to ours. As we shall see, these both commit to identifying $\ast$ with $\astN$--which we have argued is undesirable--and exhibit  further shortcomings.\footnote{We have left for a future occasion the comparison of our approach with the somewhat complex ``c-revision'' framework of  \cite{DBLP:journals/amai/Kern-Isberner04}, defined in terms of transformations of ``conditional valuation functions'', which include  probability, possibility and ranking functions as special cases.}

\subsection{Hansson}

\cite{Hansson1992-HANIDO-3} also takes a distance based approach, albeit an unconstrained one. He proposes to use the operator $\ast_{\mathrm{H}}$:

\begin{restatable}{defo}{HanssonRev}
\label{HanssonRev}
$\cbel{\Psi\ast_{\mathrm{H}} A\shortTo B}:=\bigcap \cbel{\Theta_i}$, such that the $\Theta_i$ minimise the distance to $\Psi$, subject to the constraint that $A\shortTo B\in \cbel{\Theta_i}$. 
\end{restatable}

\noindent The fate of this suggestion, of course, hinges on (i) one's view of the nature of states and (ii) the distance metric used.  But if one equates states with TPOs and measures distance by means of $d_K$, then, first of all, rational revision coincides with natural revision: $\Psi\ast  \top\shortTo B = \Psi\astN B$. Indeed:  

\begin{restatable}{prop}{closestNat}
\label{closestNat}
Let $\star$ be a revision operator that satisfies AGM. Then, if  $\preccurlyeq_{\Psi\astN A}\neq \preccurlyeq_{\Psi\star A}$, then  $d_K(\preccurlyeq_{\Psi\astN A}, \preccurlyeq_{\Psi})< d_K(\preccurlyeq_{\Psi\star A}, \preccurlyeq_{\Psi})$.
\end{restatable}

\noindent This, we take, is already not an appealing feature. Furthermore, Hansson's use of the intersection of a set of rational conditional belief sets should raise concerns, since it is well known that such intersections can fail to be rational.  As it turns out, this worry is substantiated, and his suggestion is in fact inconsistent with at least one of the AGM postulates:

\begin{restatable}{prop}{HanssonProblem}
\label{HanssonProblem}
The operator $\ast_{\mathrm{H}}$ does not satisfy
\begin{tabbing}
\=BLAHHHI \=\kill
\> \KRev{8} \> If $\neg B\notin [\Psi*A]$, then $\textrm{Cn}([\Psi*A]\cup\{B\})\subseteq$\\
\>\>$ [\Psi*A\wedge B]$\\[-0.25em]
\end{tabbing} 
\vspace{-1em}
\end{restatable}

\noindent An alternative way of aggregating the closest TPOs, which would guarantee an AGM-compliant output, would be to make use of an extension to the n-ary case of the binary TPO aggregation operator $\STQ$ of \cite{DBLP:journals/ai/BoothC19}. We leave the study of this option to those who are more enthusiastic about the prospects of natural revision.
%\subsection*{Nayal {\em et al}}
%
%Most strengthenings of DP do not give us 
%\begin{tabbing}
%\=BLAH:\=\kill
%
%\>  \>  $A \shortTo B \in \cbel{\Psi\ast A \shortto B}$   \\[-0.25em]
%
%\end{tabbing} 
%\vspace{-0.75em}
%
%\noindent \ldots with the exception of $\astL$.  \cite{DBLP:conf/ecai/NayakPFP96} simply suggest that $\Psi\ast  A\shortTo B = \Psi\astL A\shortto B$. Furthermore, on my proposal for extending $\astL$, $\Psi\astL  A\shortTo B = \Psi\astL A\shortto B$. So their suggestion satisfies the suggested properties.

\subsection{Boutilier \& Goldszmidt}

\cite{DBLP:conf/aaai/BoutilierG93} 
%\cite{boutilier-goldszmidt:1995a} \jake{only mention the second paper if we are going to discuss it!} 
offer an alternative extension of  $\astN$, which makes use of two further standard belief change operators: (i) the {\em contraction} operator $\contract$, which returns the posterior state $\Psi \contract A$ that results from an adjustment of $\Psi$ to accommodate the retraction of $A$ and (ii) the {\em expansion} operator $+$, which is similar to revision, save that consistency of the resulting beliefs needn’t be ensured.

Like ours, their proposal involves a two-stage process, this time involving a first step of contraction by $A\shortTo \neg B$, then a step of expansion by $A\shortTo B$. In the case in which $A\shortTo B$ is not initially accepted, the contraction step involves moving the minimal $\mods{A\wedge B}$ worlds down to the rank $r$ in which the minimal $\mods{A\wedge \neg B}$ worlds sit. The expansion step then has these minimal $\mods{A\wedge \neg B}$ worlds move up to a position immediately above $r$, while preserving their relations with any worlds that were strictly above or below them. Formally:

\begin{restatable}{defo}{BGcontract}
\label{BGcontract}
The {\em Boutilier-Goldszmidt contraction operator ${\contract_{\mathrm{BG}}}$} is such that
\begin{itemize}
\item[]
\begin{itemize}

\item[(1)] If $x, y\notin \min(\preccurlyeq_{\Psi}, \mods{A\wedge B})$, then $x\preccurlyeq_{\Psi {\contract_\mathrm{BG}}A\shortTo \neg B} y$ iff $x\preccurlyeq_{\Psi} y$, and  

\item[(2)] If $x\in \min(\preccurlyeq_{\Psi}, \mods{A\wedge B})$, then 

\begin{itemize}

\item[(a)] $x\preccurlyeq_{\Psi {\contract_\mathrm{BG}}A\shortTo \neg B} y$ iff $z\preccurlyeq_{\Psi} y$, for some $z\in\min(\preccurlyeq_{\Psi}, \mods{A})$, and

\item[(b)] $y\preccurlyeq_{\Psi {\contract_\mathrm{BG}}A\shortTo \neg B} x$ iff $y\preccurlyeq_{\Psi} z$, for some $z\in\min(\preccurlyeq_{\Psi}, \mods{A})$
%\footnotemark

\end{itemize}
\end{itemize}
\end{itemize}
\end{restatable}

%\footnotetext{Note that (2) gives us: If $x\in \min(\preccurlyeq_{\Psi}, \mods{A\wedge B})$ and $z\in\min(\preccurlyeq_{\Psi}, \mods{A})$, then $x\sim^\contract_{A\shortTo \neg B} z$, which given (1), is, I think, enough to determine the posterior TPO.}

%\noindent In relation to expansion:
\begin{restatable}{defo}{BGexpand}
\label{BGexpand}
The {\em Boutilier-Goldszmidt expansion operator ${+_{\mathrm{BG}}}$} is such that
\begin{itemize}
\item[]
\begin{itemize}

\item[(1)] If $x\notin \min(\preccurlyeq_{\Psi}, \mods{A\wedge \neg B})$, then $x\preccurlyeq_{\Psi {+_{\mathrm{BG}}} A\shortTo B} y$ iff $x\preccurlyeq_{\Psi} y$, and  

\item[(2)] If $x\in \min(\preccurlyeq_{\Psi}, \mods{A\wedge \neg B})$, then 
\begin{itemize}

\item[(a)] if $y\in \min(\preccurlyeq_{\Psi}, \mods{A\wedge \neg B})$, then $y\preccurlyeq_{\Psi {+_{\mathrm{BG}}} A\shortTo B} x$, and 

\item[(b)]  if $y\notin \min(\preccurlyeq_{\Psi}, \mods{A\wedge \neg B})$, then $x\preccurlyeq_{\Psi {+_{\mathrm{BG}}} A\shortTo B} y$ iff $x\preccurlyeq_{\Psi} y$ and there is no $z\in\min(\preccurlyeq_{\Psi}, \mods{A\wedge B})$ such that $y\preccurlyeq_{\Psi} z$\footnotemark{}

\end{itemize}
\end{itemize}
\end{itemize}
\end{restatable}

\footnotetext{This is not quite the original formulation, which, in view of  the informal description provided by the authors, appears to include a number of typographical errors. 
}

\noindent The  corresponding revision operator is then defined as the composition of $\contract_{\mathrm{BG}} A\shortTo \neg B$ and $+_{\mathrm{BG}} A\shortTo  B$:

\begin{restatable}{defo}{BGrevise}
\label{BGrevise}
The {\em Boutilier-Goldszmidt revision operator ${\ast_{\mathrm{BG}}}$}  is given by $\Psi \ast_{\mathrm{BG}} A\shortTo B := (\Psi \contract_{\mathrm{BG}} A\shortTo \neg B) +_{\mathrm{BG}} A\shortTo B$. 
\end{restatable}

\noindent The operation $\plusBG A\shortTo B$  bears some striking similarities to the second step in our construction of $\ast A\shortTo B$. In fact, it  {\em coincides} with it in the kind of circumstances under which it is supposed to operate, i.e. on the heels of $\contractBG A\shortTo \neg B$.

%\begin{restatable}{prop}{Coincidence}
%\label{Coincidence}
%If  $x\preccurlyeq_{\Psi \ast A\shortto B} y$ for $x\in\min(\preccurlyeq_{\Psi \ast A\shortto B}, \mods{A\wedge B})$ and $y\in\min(\preccurlyeq_{\Psi \ast A\shortto B}, \mods{A\wedge \neg B})$, then 
% $x\preccurlyeq_{\Psi \circledast A\shortTo B} y$ iff $x\preccurlyeq_{(\Psi \ast A\shortto B)\plusBG A\shortTo B} y$ \jake{cut this shit out}
%\end{restatable}

\begin{figure}[!h]
\begin{centering}
\begin{tikzpicture}[->,>=stealth']
\scalebox{1}{%
 \node[state,
  text width=2cm,        % max text width 
 ] (K) 
 {
 \scalebox{0.6}{
\begin{tabular}[c]{c|c|c}
     \multicolumn{1}{c|}{$A\wedge B$} &  $\neg A$   &   \multicolumn{1}{c}{$A\wedge \neg B$} \\
    \hline
    \rule{0pt}{2em}
     \tikzmark{T1}1\tikzmark{B1}   &    \tikzmark{T2}2\tikzmark{B2}  & \tikzmark{T3}3\tikzmark{B3} \\
     \tikzmark{T4}4\tikzmark{B4}    & \tikzmark{T5}5\tikzmark{B5} &    \\
        &    \tikzmark{T6}6\tikzmark{B6}     &  \\
         &    \tikzmark{T7}7\tikzmark{B7}      &    \\
          &       &      \tikzmark{T8}8\tikzmark{B8}      \\[1em] 
  \end{tabular}
  \DrawBox[thick, centuryred, dashed]{T4}{B4}
%        \DrawBox[thick, red]{T6}{B9}
}
};

% Next node: astN
 \node[state,       % layout (defined above)
 node distance=3.5cm,     % distance to K
  text width=2cm,        % max text width
below of=K,        % Position is to the right of K
 yshift=+0cm] (contract)    % move 3cm in y
 {
 \scalebox{0.6}{
\begin{tabular}[c]{c|c|c}
     \multicolumn{1}{c|}{$A\wedge B$} &  $\neg A$   &   \multicolumn{1}{c}{$A\wedge \neg B$} \\
    \hline
    \rule{0pt}{2em}
     \tikzmark{T1}1\tikzmark{B1}   &    \tikzmark{T2}2\tikzmark{B2}  & \tikzmark{T3}3\tikzmark{B3} \\
     & \tikzmark{T5}5\tikzmark{B5} &    \\
        &    \tikzmark{T6}6\tikzmark{B6}     &  \\
         &    \tikzmark{T7}7\tikzmark{B7}      &    \\
        \tikzmark{T4}4\tikzmark{B4}      &       &      \tikzmark{T8}8\tikzmark{B8}      \\[1em] 
  \end{tabular}
\DrawBox[thick, centuryred, dashed]{T4}{B4}
%        \DrawBox[thick, red]{T6}{B9}
}
};

% Next node: astNMod
 \node[state,       % layout (defined above)
 node distance=3.5cm,     % distance to K
  text width=2cm,        % max text width
right of=contract,        % Position is to the right of K
 yshift=+0cm] (expand)    % move 3cm in y
 {%                     % posistion relative to the center of the 'box'
 \scalebox{0.6}{
\begin{tabular}[c]{c|c|c}
     \multicolumn{1}{c|}{$A\wedge B$} &  $\neg A$   &   \multicolumn{1}{c}{$A\wedge \neg B$} \\
    \hline
    \rule{0pt}{2em}
     \tikzmark{T1}1\tikzmark{B1}   &    \tikzmark{T2}2\tikzmark{B2}  & \tikzmark{T3}3\tikzmark{B3} \\
     & \tikzmark{T5}5\tikzmark{B5} &    \\
        &    \tikzmark{T6}6\tikzmark{B6}     &  \\
         &    \tikzmark{T7}7\tikzmark{B7}      &    \\
            &       &      \tikzmark{T8}8\tikzmark{B8}      \\
        \tikzmark{T4}4\tikzmark{B4}      &       &           \\[1em] 
  \end{tabular}
\DrawBox[thick, centuryred, dashed]{T4}{B4}
%        \DrawBox[thick, red]{T6}{B9}
}
};

% Next node: astNMod
 \node[state,       % layout (defined above)
 node distance=3.5cm,     % distance to K
  text width=2cm,        % max text width
right of=K,        % Position is to the right of K
 yshift=+0cm] (us)    % move 3cm in y
 {%                     % posistion relative to the center of the 'box'
 \scalebox{0.6}{
\begin{tabular}[c]{c|c|c}
     \multicolumn{1}{c|}{$A\wedge B$} &  $\neg A$   &   \multicolumn{1}{c}{$A\wedge \neg B$} \\
    \hline
    \rule{0pt}{2em}
     \tikzmark{T1}1\tikzmark{B1}   &    \tikzmark{T2}2\tikzmark{B2}  & \tikzmark{T3}3\tikzmark{B3} \\
               &       &      \tikzmark{T8}8\tikzmark{B8}      \\
     \tikzmark{T4}4\tikzmark{B4}    & \tikzmark{T5}5\tikzmark{B5} &    \\
        &    \tikzmark{T6}6\tikzmark{B6}     &  \\
         &    \tikzmark{T7}7\tikzmark{B7}      &    \\[1em] 
  \end{tabular}
%\DrawBox[thick, centuryred, dashed]{T4}{B4}
%        \DrawBox[thick, red]{T6}{B9}
}
};

 % draw the paths and and print some Text below/above the graph
 \path (K) edge  node[anchor=south, right]
                   {
                   \scalebox{0.6}{
                   $\contract_{\mathrm{BG}}~A\shortto \neg B$
                   }
                   } (contract)
;

 \path (K.east) edge  node[anchor=north, above]
                   {
                   \scalebox{0.6}{
                   $\astN~A\shortTo B$
                   }
                   } (us)
;

 \path (contract) edge node[anchor=north,above]
                   {
                   \scalebox{0.6}{
                   $+_{\mathrm{BG}}~A\shortto B$
                   }
                   } (expand)
;
}
\end{tikzpicture}
\end{centering}

\caption{Illustrations of Boutilier \& Goldszmidt's two-step proposal for extending $\astN$, contrasted with our own. The set $\min(\preccurlyeq_{\Psi}, \mods{A\wedge B})$  is marked by a dashed box. }   
\label{fig:BG}
\end{figure}
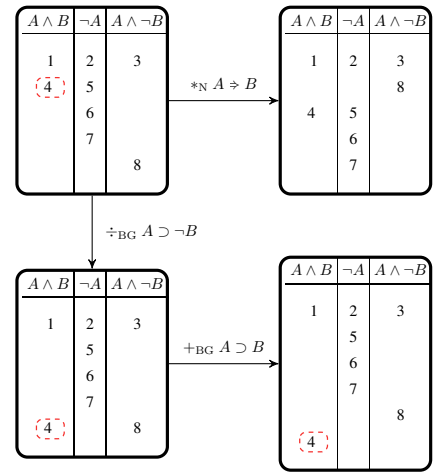

Having said that, the introduction of the contraction step means that, overall, Boutilier \& Goldszmidt's proposal quite clearly departs from the proposed extension of $\astN$ put forward in the previous section. In Figure \ref{fig:BG},  we see that it notably violates the requirement \pprincRevR{1}, according to which the ordering internal to $\mods{A\shortto B}$ should be preserved (since, although $4, 7\in \mods{A\shortto B}$  and $7\prec_{\Psi} 4$, we have $4\prec_{\Psi {\ast_{\mathrm{BG}}} A\shortTo B} 7$). This particular example is also an instance of the following feature of  their revision operator:

\begin{restatable}{prop}{ModusTollens}
\label{ModusTollens}
If $A\in \bel{\Psi}$, then $A\wedge B\in \bel{\Psi \ast_{\mathrm{BG}} A\shortTo B}$
\end{restatable}

\noindent But this is a rather questionable property: it essentially precludes reasoning by Modus Tollens (aka denying the consequent). The following example highlights the counterintuitive character of this proscription:
\begin{example}
 I believe that the light in the bathroom next door is on ($A$), because the light switch in this room is down ($\neg B$). The owner of the house tells me that, contrary to what one might expect, when  the bathroom light  is on, that means that the switch in this room is {\em up}. So I revise by $A\shortTo B$. In doing so, I maintain my belief about the state of the switch ($\neg B$) and conclude that the bathroom light is off ($\neg A$).
\end{example}

%\subsection*{Kern-Isberner}
%
% \jake{definition and diagrams} Kern-Isberner [KI01] extends a particular family of reductive revision operators ($\neq$ $\astN$, $\astR$ or $\astL$).  The extension satisfies \SRevR~and \pprincRev{1}-\pprincRev{4}.  It does not satisfy the distance-minimisation feature: the present proposal would yield a different extension 

%=====================================================

\section{Concluding comments}
\label{s:CComments}

%=====================================================

In what precedes we have offered a fresh approach to the problem of revision by conditionals, which imposes no constraints on the behaviour of the revision operator in relation to non-conditional inputs. This independence was achieved by deriving the result of a revision by a conditional from the result of a revision by its material counterpart. This approach, we have argued, satisfies a number of attractive new properties and enjoys a number of distinctive advantages over existing alternative proposals.

%One result which we would have liked to obtain is a syntactic counterpart of Theorem \ref{Representation}, in which distance \jake{???}

Having said that,  the scope of a number of results that we have established could perhaps be broadened. 

Firstly, Proposition \ref{PropertiesThree} shows that, at the level of the internal ordering of $\mods{A}$, conditional revision by $A\shortTo B$ operates like revision by $B$, for the special case of extensions of {\em elementary} revision operators. We do not know to what extent this generalises to a broader class of revision operators, such as the POI revision operators of \cite{DBLP:conf/kr/0001C18}.

Secondly, we establish in Proposition \ref{NatAndRatClosure}  that, if $\ast=\astN$ and  $A\shortTo \neg B\notin\cbel{\Psi}$, then $\cbel{\Psi\circledast A\shortTo B}=\CRat(\cbel{\Psi}\cup\{A\shortTo B\})$. This raises the following question: For $\ast = \astL$ or $\ast=\astR$, if $A\shortTo \neg B\notin\cbel{\Psi}$, do we have $\cbel{\Psi\ast A\shortTo B}=\C(\cbel{\Psi}\cup\{A\shortTo B\})$ for some suitable closure operator $\C$? 

Finally, at a number of points, we have made use of the distance metric $d_K$, noting that it was ubiquitous in the social choice literature. We are however aware of at least one alternative to this metric, proposed in \cite[Sec.~3.2]{10.2307/41681743}, which coincides with $d_K$ in the special case of linear orders. It would be interesting to assess the impact of this alternative choice on the proposal made here (another potential point of relevance concerns our assessment of Hansson's proposal, which also made use of $d_K$). 

In addition to the question of the generalisability of certain results, we note that there is an extensive literature on a related  issue for models of graded, rather than categorical, belief (esp.~probabilistic models): how to update one's degrees of belief on information specifying a particular conditional degree of belief or presented in the form of a natural language indicative conditional. A natural approach here  is to move to the posterior distribution that is ``closest'' to the prior one, on some appropriate distance measure, subject of the relevant informational constraint.  However, an apparent issue with the use of the popular cross-entropy measure was presented in the classic Judy Benjamin example of \cite{vanFraassen1981-VANAPF}, with similar observations being made in relation to two further measures in \cite{vanFraassen1986-VANAPF-9}. For further discussions, see \cite{doi:10.1111/j.1468-0017.2012.01443.x}, \cite{Douven2011-DOUAPA},  \cite{Douven2011-ROMANR},  \cite{EvaForthcoming-EVALFC}, and  \cite{Grove:1997:PUC:2074226.2074251}. An examination of potential points of contact with the present  work would be interesting to pursue.

%
%
%
%\vfill
%
%\pagebreak

\section*{Acknowledgements}

This research was supported by the Australian Government through an Australian Research Council  Future Fellowship (project number FT160100092) awarded to Jake Chandler.

%=====================================================

\section*{Appendix}\label{s:appendix}

%=====================================================

\vspace{1em}

\RamseyMat*

\begin{myproof}
%Assume that $A\shortto B\notin\bel{\Psi}$. By AGM, there exists $x\in\min(\preccurlyeq_{\Psi}, W)\cap\mods{A\wedge \neg B}$. Since $x\in \min(\preccurlyeq_{\Psi}, W)\cap\mods{A}$, we have $x\in\min(\preccurlyeq_{\Psi}, \mods{A})$. But, by \RT~and AGM, $A\shortTo B\in\cbel{\Psi}$ only if $\min(\preccurlyeq_{\Psi}, \mods{A})\subseteq \mods{B}$. Hence, since $x\in\mods{\neg B}$, it follows that $A\shortTo B\notin\cbel{\Psi}$.
Regarding (b) : Let $W=\{w,x,y,z\}$, with $w, x, y$ and $z$ respectively in $\mods{A\wedge B}$, $\mods{A\wedge \neg B}$, $\mods{\neg A\wedge B}$ and $\mods{\neg A\wedge \neg B}$. Let $\preccurlyeq_{\Psi}$ be given by $z\prec_{\Psi} \{w, x, y\}$.  Then $A\shortto B\in\bel{\Psi}$, but $A\shortTo B\notin\cbel{\Psi}$.
\end{myproof}

\vspace{1em}

%=====================================================

\OnlyLex*

\begin{myproof}
Given AGM, \CRevR{1}, \CRevR{2}, and $(\mathrm{Eq}^\ast_\preccurlyeq)$, $\astL$ is characterised by the `Recalcitrance' property
\begin{tabbing}
\=BLAHI: \=\kill
\>  (Rec$^\ast$) \>  If $A \wedge B$ is consistent, then $A \in \bel{(\Psi \ast A) \ast B}$. \\[-0.25em]
\end{tabbing} 
\vspace{-1em}
\noindent It will suffice to show that this property is entailed by:
\begin{tabbing}
\=BLAHI: \=\kill
\>  (1) \>  If $A \wedge B$ is consistent, then $B \in$\\
\>\> $ \bel{(\Psi \ast A\shortto B) \ast A}$ \\[-0.25em]
\end{tabbing} 
\vspace{-1em}
\noindent So let $A\wedge B$ be consistent. Since $A\equiv (A\shortto B)\shortto A$, it suffices, by (Eq$^\ast$)  to show $A \in \bel{(\Psi \ast (A\shortto B)\shortto A) \ast B}$. We know that for any AGM operator $\ast'$ and state $\Psi'$, $A\in\bel{(\Psi'\ast' B)}$ iff $A\in\bel{(\Psi'\ast' A\shortto B)}$. Hence it suffices to show $A \in \bel{(\Psi \ast (A\shortto B)\shortto A) \ast A\shortto B}$. Since $A\wedge B \equiv ((A\shortto B)\shortto A)\wedge ( A\shortto B)$ is consistent, we can apply (1) to recover the required result.
\end{myproof}

\vspace{1em}

%%=====================================================

\SyntacticPprincOne*

\begin{myproof}
\begin{itemize}

\item[{\bf (i)}] From \pprincRevR{1} to \pprincRev{1}: Assume that $A\shortto B\in\Cn(C)$. We prove each inclusion in turn
\begin{itemize}

\item[-] Regarding $\bel{(\Psi\ast A\shortto B)\ast C}\subseteq \bel{(\Psi\ast A\shortTo B)\ast C}$: Assume that $x\in\min(\preccurlyeq_{(\Psi\ast A\shortTo B)\ast C}, W)$ but, for reductio,   $x\notin\min(\preccurlyeq_{(\Psi\ast A\shortto B)\ast C}, W)$. From the former, by Success, $x\in \mods{C}$ and so, by the fact that $A\shortto B\in\Cn(C)$, it follows that $x\in \mods{A\shortto B}$. Now, by Success and $A\shortto B\in\Cn(C)$ we also have $\min(\preccurlyeq_{(\Psi\ast A\shortto B)\ast C}, W)\subseteq\mods{A\shortto B}$. Hence, since $x\notin\min(\preccurlyeq_{(\Psi\ast A\shortto B)\ast C}, W)$, there exists $y\in \mods{A\shortto B}$ such that $y\prec_{(\Psi\ast A\shortto B)\ast C} x$. But  $x\in\min(\preccurlyeq_{(\Psi\ast A\shortTo B)\ast C}, W)$ and so $x\preccurlyeq_{(\Psi\ast A\shortTo B)\ast C} y$. This contradicts  \pprincRevR{1}. Hence $x\in\min(\preccurlyeq_{(\Psi\ast A\shortto B)\ast C}, W)$, as required.

\item[-] Regarding $\bel{(\Psi\ast A\shortTo B)\ast C}\subseteq \bel{(\Psi\ast A\shortto B)\ast C}$: Similar to the item immediately above.

\end{itemize}
\item[{\bf (ii)}] From \pprincRev{1} to \pprincRevR{1}: Assume that $x, y \in \mods{A\shortto B}$. We prove each direction of the biconditional in turn
\begin{itemize}

\item[-] From $x \preccurlyeq_{\Psi\ast A\shortTo B}y$ to $x \preccurlyeq_{\Psi\ast A\shortto B}y$: Assume $x \preccurlyeq_{\Psi\ast A\shortTo B}y$, but, for reductio, $y \prec_{\Psi\ast A\shortto B}x$. Then, by Success, $\neg x\in\bel{(\Psi\ast A\shortto B)\ast x\vee y}$ but $\neg x\notin\bel{(\Psi\ast A\shortTo B)\ast x\vee y}$, contradicting  \pprincRev{1}. So  $x \preccurlyeq_{\Psi\ast A\shortto B}y$, as required.

\item[-] From $x \preccurlyeq_{\Psi\ast A\shortto B}y$ to $x \preccurlyeq_{\Psi\ast A\shortTo B}y$: Similar to the item immediately above.

\end{itemize}
\end{itemize}
\end{myproof}

\vspace{1em}

%=====================================================

\Representation*

\begin{myproof}
We prove the result in three lemmas. The first of these is the following:
\begin{restatable}{lem}{TheBasics}
\label{lem:TheBasics}
For any function $\ast$ from $\mathbb{S} \times L$ to $\mathbb{S}$, $\circledast$ satisfies \SRevR~and \pprincRevR{1}. 
\end{restatable}

\noindent We note that, given the definition of $\astL$ in Definition \ref{elemdef},  Definition \ref{MyProposal} is equivalent to:
\begin{itemize}

\item[] $x \preccurlyeq_{\Psi \circledast A\shortTo B}  y$ iff 
\begin{itemize}

\item[(1)]  $x\in D(\preccurlyeq_{\Psi\ast A\shortto B}, A\wedge B)\cap\mods{A\shortto B}$ and $y\notin D(\preccurlyeq_{\Psi\ast A\shortto B}, A\wedge B)\cap\mods{A\shortto B}$, or 

\item[(2)]  ($x\in D(\preccurlyeq_{\Psi\ast A\shortto B}, A\wedge B)\cap\mods{A\shortto B}$ iff $y\in D(\preccurlyeq_{\Psi\ast A\shortto B}, A\wedge B)\cap\mods{A\shortto B}$) and $x \preccurlyeq_{\Psi\ast A\shortto B} y$

\end{itemize}
\end{itemize}
We now show that \SRevR~is satisfied. By definition, $D(\preccurlyeq_{\Psi\ast A\shortto B}, A\wedge B)\cap\mods{A\wedge B}\neq \varnothing$. So  let $x$ be such that  $x\in D(\preccurlyeq_{\Psi\ast A\shortto B}, A\wedge B)\cap\mods{A\wedge B}$. Now consider $y\in \mods{A\wedge\neg B}$. By Definition \ref{MyProposal}, $x\prec_{\Psi\circledast A\shortTo B}y$, and hence $\min(\preccurlyeq_{\Psi\circledast A\shortTo B}, \mods{A})\subseteq \mods{B}$, as required. 

Regarding \pprincRevR{1}:  Assume $x, y \in \mods{A\shortto B}$. For the left to right  direction, assume that $x\preccurlyeq_{\Psi\circledast A\shortTo B}y$. From this one of either (1) or (2) holds. (2) immediately entails that $x\preccurlyeq_{\Psi\ast A\shortto B}y$. (1) gives us: $x\in D(\preccurlyeq_{\Psi\ast A\shortto B}, A\wedge B)$ and $y \notin D(\preccurlyeq_{\Psi\ast A\shortto B}, A\wedge B)$. $x\preccurlyeq_{\Psi\ast A\shortto B}y$ then follows from the definition of $D(\preccurlyeq_{\Psi\ast A\shortto B}, A\wedge B)$. For the right to left  direction, assume that $x\preccurlyeq_{\Psi\ast A\shortto B}y$. If $x, y\in D(\preccurlyeq_{\Psi\ast A\shortto B}, A\wedge B)$ or  $x, y\notin D(\preccurlyeq_{\Psi\ast A\shortto B}, A\wedge B)$, then we obtain $x\preccurlyeq_{\Psi\circledast A\shortTo B}y$ by (1). So assume that one of either $x$ or $y$ is in $D(\preccurlyeq_{\Psi\ast A\shortto B}, A\wedge B)$, while the other is not. From $x\preccurlyeq_{\Psi\ast A\shortto B}y$, it must be the case that  $x\in D(\preccurlyeq_{\Psi\ast A\shortto B}, A\wedge B)$ and $y \notin D(\preccurlyeq_{\Psi\ast A\shortto B}, A\wedge B)$. From (2), we then again recover $x\preccurlyeq_{\Psi\circledast A\shortTo B}y$.

Our next lemma states a general fact about lexicographic combinations of ordered pairs of TPOs, defined by

\begin{restatable}{defo}{lexcombo}
\label{lexcombo}
The {\em lexicographic combination} $\mathrm{lex}(\preccurlyeq_1, \preccurlyeq_2)$ of two TPOs $\preccurlyeq_1$ and $\preccurlyeq_2$ is given by the TPO $\preccurlyeq$ such that $x\preccurlyeq y$ iff  (i) $x \prec_1 y$ or (ii) $x \sim_1 y$ and $x \prec_2 y$
\end{restatable}

\noindent It is given as follows:

\begin{restatable}{lem}{Lexresult}
\label{lem:Lexresult}
\noindent Let $\preccurlyeq_{1}$, $\preccurlyeq_{2}$ be two given TPOs and let $X(\preccurlyeq_{2}) =\{\preccurlyeq\mid \preccurlyeq\mbox{ is a TPO s.t.~}\preccurlyeq\subseteq \preccurlyeq_{2}\}$. Then the TPO in $X(\preccurlyeq_{2})$ that minimises the distance $d_K$ to $\preccurlyeq_{1}$ is $\mathrm{lex}(\preccurlyeq_1, \preccurlyeq_2)$.  
\end{restatable}

\noindent Let $\preccurlyeq'= \mathrm{lex}(\preccurlyeq_1, \preccurlyeq_2)$. First we need to check that $\preccurlyeq'\in X(\preccurlyeq_2)$, i.e.~$\preccurlyeq'\subseteq\preccurlyeq_2$. But this is clear from Definition \ref{lexcombo}. It remains to be shown that $d_K(\preccurlyeq_1, \preccurlyeq') \leq d_K(\preccurlyeq_1, \preccurlyeq'')$ for all $\preccurlyeq''\in X(\preccurlyeq_2)$. To see this, we first reformulate $d_K$.

\begin{restatable}{defo}{hardsoft}
\label{hardsoft}
A {\em hard conflict} between $\preccurlyeq_1, \preccurlyeq'$ is a 2-element set $\{x, y\}$ s.t.~$x\prec' y$ and $y\prec_{1} x$. Let $\mathrm{Hard}(\preccurlyeq_1, \preccurlyeq')$ denote the set of such hard conflicts.

A {\em soft conflict} between $\preccurlyeq_1, \preccurlyeq'$ is a 2-element set $\{x, y\}$ s.t.~either (i) $x\prec' y$ and $x\sim_{1} y$ or (ii) $x\prec_1 y$ and $x\sim' y$. Let $\mathrm{Soft}(\preccurlyeq_1, \preccurlyeq')$ denote the set of such soft conflicts. 
\end{restatable}

\noindent So $d_K(\preccurlyeq_1, \preccurlyeq') = 2\times \lvert \mathrm{Hard}(\preccurlyeq_1, \preccurlyeq')\rvert + \lvert \mathrm{Soft}(\preccurlyeq_1, \preccurlyeq')\rvert$ and similarly for $d_K(\preccurlyeq_1, \preccurlyeq'')$. Hence, to show that $d_K(\preccurlyeq_1, \preccurlyeq') < d_K(\preccurlyeq_1, \preccurlyeq'')$ when $\preccurlyeq'\neq \preccurlyeq''$, it  suffices to prove
\begin{itemize}
\item[] (1) $\mathrm{Hard}(\preccurlyeq_1, \preccurlyeq')\subseteq \mathrm{Hard}(\preccurlyeq_1, \preccurlyeq'')$ 

\item[] (2) $\mathrm{Soft}(\preccurlyeq_1, \preccurlyeq')\subseteq \mathrm{Soft}(\preccurlyeq_1, \preccurlyeq'')$ 
\end{itemize}
Regarding (1): Let $\{x, y\}\in \mathrm{Hard}(\preccurlyeq_1, \preccurlyeq')$, i.e. $x\prec_1 y$ and $y\prec' x$. We must show $y\prec'' x$. By definition of $\preccurlyeq'= \mathrm{lex}(\preccurlyeq_1, \preccurlyeq_2)$, we have, from $y\prec' x$, (i) $y\prec_2 x$ or (ii) $y\sim_2 x$ and $y\prec_1 x$. We cannot have $y\prec_1 x$, since we already have $x\prec_1 y$. So $y\prec_2 x$. Hence, since $\preccurlyeq''\in X(\preccurlyeq_2)$, i.e. $\preccurlyeq''\subseteq\preccurlyeq_2$, we have $y\prec' x$ as well, as required.

Regarding (2): Let $\{x, y\}\in \mathrm{Soft}(\preccurlyeq_1, \preccurlyeq')$. We then have two cases to consider:
\begin{itemize}

\item[-] $x\sim_1 y$ and $x\prec' y$: From $x\prec' y$, we get either (i) $x\prec_2 y$ or (ii) $x\sim_2 y$ and $x\prec_1 y$. The latter cannot occur, since we assume  $x\sim_1 y$. Hence $x\prec_2 y$. Since $\preccurlyeq''\subseteq \preccurlyeq_2$, we also then have $x\prec'' y$, so $\{x,y\}\in\mathrm{Soft}(\preccurlyeq_1, \preccurlyeq'')$, as required.

\item[-] $x\sim' y$ and $x\prec_1 y$: Impossible, since $x\sim' y$ entails that both $x\sim_1 y$  and $x\sim_2 y$ but we assume $x\prec_1 y$.

\end{itemize}
We now show that:
\begin{restatable}{lem}{Subsetresult}
\label{lem:Subsetresult}
For any $\preccurlyeq'$ satisfying \SRevR~and \pprincRevR{1}, we must have $\preccurlyeq'\subseteq \preccurlyeq_{D}$, where $\preccurlyeq_D$ is defined as follows:
\begin{itemize}

\item[] $x \preccurlyeq_D y$ iff $x\in D(\preccurlyeq_{\Psi\ast A\shortto B}, A\wedge B)\cap\mods{A\shortto  B}$ or $y\notin D(\preccurlyeq_{\Psi\ast A\shortto B}, A\wedge B)\cap\mods{A\shortto  B}$

\end{itemize}
\end{restatable}

\noindent Let $\preccurlyeq'$ satisfy \SRevR~and \pprincRevR{1}. Suppose $y\preccurlyeq_D x$. We must show that $y\prec' x$. From $y\preccurlyeq_D x$, by definition of $\preccurlyeq_D$,  we have $x\notin D(\preccurlyeq_{\Psi\ast A\shortto B}, A\wedge B)\cap\mods{A\shortto  B}$ and  $y\in D(\preccurlyeq_{\Psi\ast A\shortto B}, A\wedge B)\cap\mods{A\shortto  B}$. From $y\in D(\preccurlyeq_{\Psi\ast A\shortto B}, A\wedge B)$, we have $y\preccurlyeq_{\Psi\ast A\shortto B}z$, where $z\in\min(\preccurlyeq_{\Psi\ast A\shortto B}, \mods{A\wedge B})$. We now consider two cases:
\begin{itemize}

\item[-] $x\in\mods{A\shortto B}$: Then, since $x\notin D(\preccurlyeq_{\Psi\ast A\shortto B}, A\wedge B)\cap\mods{A\shortto  B}$, we have $x\notin D(\preccurlyeq_{\Psi\ast A\shortto B}, A\wedge B)$ and so, form this and $y\in D(\preccurlyeq_{\Psi\ast A\shortto B}, A\wedge B)$, we get $y\preccurlyeq_{\Psi\ast A\shortto B}x$. Since $x, y \in\mods{A\shortto B}$, we get from this $y\prec' x$ by \pprincRevR{1}, as required.

\item[-] $x\in\mods{A\wedge\neg B}$: By  \SRevR, we know that $u\prec' x$ for some $u\in\mods{A\wedge B}$. By the minimality of $z$, we know $z\preccurlyeq_{\Psi\ast A\shortto B} u$. Hence  $y\preccurlyeq_{\Psi\ast A\shortto B} u$. Then, by \pprincRevR{1}, we recover   $y\preccurlyeq' u$ and so  $y\prec' x$, as required.

\end{itemize}
Putting together Lemmas \ref{lem:TheBasics}, \ref{lem:Lexresult} and \ref{lem:Subsetresult} yields the proof of the theorem. Lemma \ref{lem:TheBasics} tells us that $\preccurlyeq_{\Psi\circledast A\shortTo B}$ satisfies \SRevR~and  \pprincRevR{1}. Next, note that Definition \ref{MyProposal} can be equivalently presented in terms of a lexicographic combination, so that $\preccurlyeq_{\Psi\circledast A\shortTo B} = \mathrm{lex}(\preccurlyeq_D, \preccurlyeq_{\Psi\ast A\shortto B})$.

 In view of this, we can see that, by Lemma \ref{lem:Lexresult}, $\preccurlyeq_{\Psi\circledast A\shortTo B}$ minimises $d_K$ to $\preccurlyeq_{\Psi\ast A\shortto B}$ among all TPOs $\preccurlyeq$ s.t.~$\preccurlyeq\subseteq\preccurlyeq_D$. Finally, since all $\preccurlyeq$ that satisfy \SRevR~and  \pprincRevR{1} are such  that $\preccurlyeq\subseteq\preccurlyeq_D$, by Lemma \ref{lem:Subsetresult}, $\preccurlyeq_{\Psi\circledast A\shortTo B}$ must also minimise $d_K$ among all TPOs satisfying \SRevR~and  \pprincRevR{1} .
\end{myproof}

\vspace{1em}

%=====================================================

\SyntacticMinimality*

\begin{myproof}
We will prove the equivalent statement:  if $\bel{(\Psi\circledast A\shortTo B)\circledast C}\neq \bel{(\Psi\circledast A\shortto B)\circledast C}$, then $\bel{(\Psi\ast' A\shortTo B)\ast' C}\neq \bel{(\Psi\ast' A\shortto B)\ast' C}$.

Suppose $\bel{(\Psi\circledast A\shortTo B)\circledast C}\neq \bel{(\Psi\circledast A\shortto B)\circledast C}$. Then $\min(\preccurlyeq_{\Psi\circledast A\shortTo B}, \mods{C})\neq \min(\preccurlyeq_{\Psi\circledast A\shortto B}, \mods{C})$. So we have two cases to consider, corresponding to the failures of each direction of subset inclusion.

Assume $\min(\preccurlyeq_{\Psi\circledast A\shortTo B}, \mods{C})\nsubseteq  \min(\preccurlyeq_{\Psi\ast A\shortto B}, \mods{C})$. Let $x\in \min(\preccurlyeq_{\Psi\circledast A\shortTo B}, \mods{C}) - \min(\preccurlyeq_{\Psi\ast A\shortto B}, \mods{C})$. Let $y\in \min(\preccurlyeq_{\Psi\ast A\shortto B}, \mods{C})$. Then $x \preccurlyeq_{\Psi\circledast A\shortTo B}  y$ (by minimality of $x$) and $y \preccurlyeq_{\Psi\ast A\shortto B}  x$. Since $x \preccurlyeq_{\Psi\circledast A\shortTo B}  y$ , we know that either $x \sim_{\Psi\circledast A\shortTo B}  y$ or $x \prec_{\Psi\circledast A\shortTo B}  y$. By our construction of $\circledast$, it is not possible to have both $x \sim_{\Psi\circledast A\shortTo B}  y$ and $y \prec_{\Psi\ast A\shortto B}  x$. So we must have $x \prec_{\Psi\circledast A\shortTo B}  y$ and hence $\{x, y \}\in \mathrm{Hard}(\preccurlyeq_{\Psi\circledast A\shortTo B}, \preccurlyeq_{\Psi\ast A\shortto B})$. Now, since we assume $\ast'$ to satisfy \SRevR~and  \pprincRevR{1}, we know from our proof of Theorem \ref{Representation} that $\mathrm{Hard}(\preccurlyeq_{\Psi\circledast A\shortTo B}, \preccurlyeq_{\Psi\ast A\shortto B})\subseteq \mathrm{Hard}(\preccurlyeq_{\Psi\ast' A\shortTo B}, \preccurlyeq_{\Psi\ast A\shortto B})$. So $\{x, y \}\in \mathrm{Hard}(\preccurlyeq_{\Psi\ast' A\shortTo B}, \preccurlyeq_{\Psi\ast A\shortto B})$ and hence $x \prec_{\Psi\ast' A\shortTo B} y$. So we have  $y\in \min(\preccurlyeq_{\Psi\ast' A\shortto B}, \mods{C})=\min(\preccurlyeq_{\Psi\ast A\shortto B}, \mods{C})$ (since  $\ast'$ extends $\ast$, i.e.~agrees with it on unconditional revisions) but $y\notin \min(\preccurlyeq_{\Psi\ast' A\shortto B}, \mods{C})$. So  $\min(\preccurlyeq_{\Psi\ast' A\shortto B}, \mods{C})\neq \min(\preccurlyeq_{\Psi\ast' A\shortTo B}, \mods{C})$, i.e.~ $\bel{(\Psi\ast' A\shortTo B)\ast' C}\neq \bel{(\Psi\ast' A\shortto B)\ast' C}$, as required

Assume $\min(\preccurlyeq_{\Psi\ast A\shortto B}, \mods{C})\nsubseteq  \min(\preccurlyeq_{\Psi\circledast A\shortTo B}, \mods{C})$. Let $x\in \min(\preccurlyeq_{\Psi\ast A\shortto B}, \mods{C}) - \min(\preccurlyeq_{\Psi\circledast A\shortTo B}, \mods{C})$. Let $y\in \min(\preccurlyeq_{\Psi\circledast A\shortTo B}, \mods{C})$. Then $y\prec_{\Psi\circledast A\shortTo B} x$ and $x \preccurlyeq_{\Psi\ast A\shortto B} y$, from which $x \prec_{\Psi\ast A\shortto B} y$ or  $x \sim_{\Psi\ast A\shortto B} y$. 

If  $x \prec_{\Psi\ast A\shortto B} y$, then $\{x, y \}\in \mathrm{Hard}(\preccurlyeq_{\Psi\circledast A\shortTo B}, \preccurlyeq_{\Psi\ast A\shortto B})$. Since, as noted above, $\mathrm{Hard}(\preccurlyeq_{\Psi\circledast A\shortTo B}, \preccurlyeq_{\Psi\ast A\shortto B})\subseteq \mathrm{Hard}(\preccurlyeq_{\Psi\ast' A\shortTo B}, \preccurlyeq_{\Psi\ast A\shortto B})$, we get $y \prec_{\Psi\ast' A\shortTo B} x$ and so $x\in \min(\preccurlyeq_{\Psi\ast' A\shortto B}, \mods{C}) - \min(\preccurlyeq_{\Psi\ast' A\shortTo B}, \mods{C})$, i.e.~$\bel{(\Psi\ast' A\shortTo B)\ast' C}\neq \bel{(\Psi\ast' A\shortto B)\ast' C}$, as required.

If  $x \sim_{\Psi\ast A\shortto B} y$, then $\{x, y \}\in \mathrm{Soft}(\preccurlyeq_{\Psi\circledast A\shortTo B}, \preccurlyeq_{\Psi\ast A\shortto B})$.  Since, $\mathrm{Soft}(\preccurlyeq_{\Psi\circledast A\shortTo B}, \preccurlyeq_{\Psi\ast A\shortto B})\subseteq \mathrm{Soft}(\preccurlyeq_{\Psi\ast' A\shortTo B}, \preccurlyeq_{\Psi\ast A\shortto B})$ (again, see proof of Theorem \ref{Representation}), we have  $\{x, y \}\in \mathrm{Soft}(\preccurlyeq_{\Psi\ast' A\shortTo B}, \preccurlyeq_{\Psi\ast A\shortto B})$, so either $x \prec_{\Psi\ast' A\shortTo B} y$ or $y \prec_{\Psi\ast' A\shortTo B} x$. In the latter case, we get $\bel{(\Psi\ast' A\shortTo B)\ast' C}\neq \bel{(\Psi\ast' A\shortto B)\ast' C}$ as above. In the former case, we deduce $y\notin \min(\preccurlyeq_{\Psi\ast' A\shortTo B}, \mods{C})$. But from $x\sim_{\Psi\ast A\shortto B} y$ and $x\in \min(\preccurlyeq_{\Psi\ast A\shortto B}, \mods{C})$ we must have  $y\in \min(\preccurlyeq_{\Psi\ast A\shortto B}, \mods{C})$. Hence $y\in \min(\preccurlyeq_{\Psi\ast A\shortto B}, \mods{C})-\min(\preccurlyeq_{\Psi\ast' A\shortTo B}, \mods{C})$, so $\min(\preccurlyeq_{\Psi\ast A\shortto B}, \mods{C})\neq \min(\preccurlyeq_{\Psi\ast' A\shortTo B}, \mods{C})$, which gives again $\bel{(\Psi\ast' A\shortTo B)\ast' C}\neq \bel{(\Psi\ast' A\shortto B)\ast' C}$, as required. 
\end{myproof}

\vspace{1em}

%=====================================================

%\AnotherRepresentationTh*

\RetSoundness*

\begin{myproof}
Given the definition of $\astL$ in Definition \ref{elemdef},  if we assume $\ast=\circledast$, then Definition \ref{MyProposal} tells us that:
\begin{itemize}

\item[] $x \preccurlyeq_{\Psi \ast A\shortTo B}  y$ iff
\begin{itemize}

\item[(1)]   (i) $x\in D(\preccurlyeq_{\Psi\ast A\shortto B}, A\wedge B)\cap\mods{A\shortto B}$ and 

(ii)  $y\notin D(\preccurlyeq_{\Psi\ast A\shortto B}, A\wedge B)\cap\mods{A\shortto B}$, 

\end{itemize}
or 
\begin{itemize}

\item[(2)]   (i) ($x\in D(\preccurlyeq_{\Psi\ast A\shortto B}, A\wedge B)\cap\mods{A\shortto B}$ iff $y\in D(\preccurlyeq_{\Psi\ast A\shortto B}, A\wedge B)\cap\mods{A\shortto B}$) and  

(ii) $x \preccurlyeq_{\Psi\ast A\shortto B} y$

\end{itemize}
\end{itemize}

\noindent Regarding  \pprincRevR{2}: Assume $x, y \in \mods{A\wedge \neg B}$. For the left to right  direction, assume that $x\preccurlyeq_{\Psi\ast A\shortTo B}y$. From this one of either (1) or (2) holds. Since, $x, y \in \mods{A\wedge \neg B}$, it must be the case that (2). But this immediately entails that $x\preccurlyeq_{\Psi\ast A\shortto B}y$. The proof of the right to left direction is entirely analogous to the one given immediately above. 

Regarding \pprincRevR{3}: Assume $x\in \mods{A\shortto B}$, $y\in \mods{A\wedge \neg B}$, and $x \prec_{\Psi\ast A\shortto B} y$. We want to show that $x \prec_{\Psi\ast A\shortTo B} y$, i.e.~ that both
\begin{itemize}

\item[(3)]   (i) $y\notin D(\preccurlyeq_{\Psi\ast A\shortto B}, A\wedge B)$ or 

(ii) $y\in\mods{A\wedge \neg B}$ or 

 (iii) $x\in D(\preccurlyeq_{\Psi\ast A\shortto B}, A\wedge B)\cap\mods{A\shortto B}$, 
 
\end{itemize}
and 
\begin{itemize}

\item[(4)]   (i) $y\in D(\preccurlyeq_{\Psi\ast A\shortto B}, A\wedge B)\cap\mods{A\shortto B}$ iff $x\notin D(\preccurlyeq_{\Psi\ast A\shortto B}, A\wedge B)\cap\mods{A\shortto B}$ or 

(ii) $x \prec_{\Psi\ast A\shortto B} y$
 
\end{itemize}
Our assumption that $y\in \mods{A\wedge \neg B}$ gets us (3), while $x \prec_{\Psi\ast A\shortto B} y$ gets us (4). 

Regarding  \pprincRevR{4}: Assume $x\in \mods{A\shortto B}$, $y\in \mods{A\wedge \neg B}$, and $x \preccurlyeq_{\Psi\ast A\shortto B} y$. We want to establish the disjunction of (1) and (2). If $x\in D(\preccurlyeq_{\Psi\ast A\shortto B}, A\wedge B)$, then, since  $x\in \mods{A\shortto B}$ and $y\in \mods{A\wedge \neg B}$, we have (1) and we are done. So assume $x\notin D(\preccurlyeq_{\Psi\ast A\shortto B}, A\wedge B)$ and hence, by $x \preccurlyeq_{\Psi\ast A\shortto B} y$, $x, y\notin D(\preccurlyeq_{\Psi\ast A\shortto B}, A\wedge B)$. (2) then holds by virtue of  this and the fact that $x \preccurlyeq_{\Psi\ast A\shortto B} y$. 
\end{myproof}

\vspace{1em}

%%=====================================================

\PropertiesTwo*

\begin{myproof}
The equivalence and entailments are obvious. Regarding the failure of entailment from \pprincRevR{3'} and  \pprincRevR{4'} to \pprincRevR{3} and  \pprincRevR{4}, see the  countermodel in Figure \ref{fig:PrimeCM}.

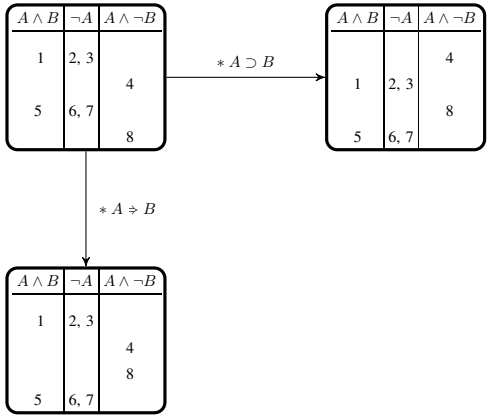
\begin{figure}[!h]
\begin{centering}
\begin{tikzpicture}[->,>=stealth']
\scalebox{1}{%
 \node[state, 
   text width=2.1cm] (K) 
 {
 \scalebox{0.6}{
\begin{tabular}[c]{c|c|c}
     \multicolumn{1}{c|}{$A\wedge B$} &  $\neg A$   &   \multicolumn{1}{c}{$A\wedge \neg B$} \\
    \hline
    \rule{0pt}{2em}
     1  &  2, 3   &   \\
       &     &  4  \\
     5  &  6, 7   &   \\
       &     &  8 \\
  \end{tabular}
}
};

% Next node: astR
 \node[state,       % layout (defined above)
 node distance=4.25cm,     % distance to K
  text width=2.1cm,
right of=K,        % Position is to the right of K
 yshift=+0cm] (astR)    % move 3cm in y
 {%                     % posistion relative to the center of the 'box'
  \scalebox{0.6}{
\begin{tabular}[c]{c|c|c}
     \multicolumn{1}{c|}{$A\wedge B$} &  $\neg A$   &   \multicolumn{1}{c}{$A\wedge \neg B$} \\
    \hline
    \rule{0pt}{2em}
       &     &  4  \\
     1  &  2, 3   &   \\
       &     &  8 \\
     5  &  6, 7   &   \\
  \end{tabular}
}
};

 \node[state,       % layout (defined above)
 node distance=3.5cm,     % distance to K
  text width=2.1cm,
below of=K,        % Position is to the right of K
 yshift=+0cm] (astW)    % move 3cm in y
 {%                     % posistion relative to the center of the 'box'
  \scalebox{0.6}{
\begin{tabular}[c]{c|c|c}
     \multicolumn{1}{c|}{$A\wedge B$} &  $\neg A$   &   \multicolumn{1}{c}{$A\wedge \neg B$} \\
    \hline
    \rule{0pt}{2em}
     1  &  2, 3   &   \\
       &     &  4  \\
       &     &  8 \\
     5  &  6, 7   &   \\
  \end{tabular}
}
};

 % draw the paths and and print some Text below/above the graph
 \path (K) edge  node[anchor=north, above]
                   {
                   \scalebox{0.6}{
                   $\ast~A\shortto B$
                   }
                   } (astR)
;

 % draw the paths and and print some Text below/above the graph
 \path (K) edge  node[anchor=west]
                   {
                   \scalebox{0.6}{
                   $\ast~A\shortTo B$
                   }
                   } (astW)
;

}
\end{tikzpicture}

\caption{Countermodel for proof of Proposition \ref{PropertiesTwo}}
\label{fig:PrimeCM}
\end{centering}
\end{figure}

There, we see that the DP postulates, as well as \SRevR, \pprincRevR{1}, \pprincRevR{2}, \pprincRevR{3'} and  \pprincRevR{4'} are satisfied. However, neither \pprincRevR{3} nor  \pprincRevR{4} are satisfied because, notably, $1 \prec_{\Psi\ast A\shortto B} 4$, but $4 \prec_{\Psi\ast A\shortTo B} 1$.
\end{myproof}

\vspace{1em}

%%=====================================================

\SyntacticPprinc*

\begin{myproof}
The derivation of the equivalences is straightforward and resembles the well known derivation of the equivalences between \CRevR{1}--\CRevR{4} and their syntactic counterparts:

\begin{itemize}

\item[{\bf (a)}] \pprincRevR{2} and \pprincRev{2}: Similar to (a) above.

\item[{\bf (b)}] \pprincRevR{3} and \pprincRev{3}:
\begin{itemize}

\item[{\bf (i)}] From \pprincRevR{3} to \pprincRev{3}: Assume that $A\rightarrow B\in\bel{(\Psi\ast A\rightarrow B)\ast C}$, but, for reductio, $A\rightarrow B\notin\bel{(\Psi\ast A\Rightarrow B)\ast C}$. Then there exists $y\in\min(\preccurlyeq_{(\Psi\ast A\shortTo B)\ast C}, W)\cap \mods{A\wedge \neg B}$. Let $x$ be in $\min(\preccurlyeq_{(\Psi\ast A\shortto B)\ast C}, W)\cap \mods{A\shortto B}$ (we know that such an $x$ exists, since $\min(\preccurlyeq_{(\Psi\ast A\shortto B)\ast C}, W)\subseteq \mods{A\shortto B}$). Since $y\in \mods{A\wedge \neg B}$ and $\min(\preccurlyeq_{(\Psi\ast A\shortto B)\ast C}, W)\subseteq \mods{A\shortto B}$, we have $x \prec_{\Psi\ast A\shortto B}y$. As $x\in  \mods{A\shortto B}$ and $y\in \mods{A\wedge \neg B}$, we can then apply \pprincRevR{3} to recover $x \prec_{\Psi\ast A\shortTo B}y$. But this contradicts $y\in\min(\preccurlyeq_{(\Psi\ast A\shortTo B)\ast C}, W)$. Hence $A\rightarrow B\in\bel{(\Psi\ast A\Rightarrow B)\ast C}$, as required.

\item[{\bf (ii)}] From \pprincRev{3} to \pprincRevR{3}: Assume $x\in \mods{A\shortto B}$, $y\in \mods{A\wedge \neg B}$, and $x \prec_{\Psi\ast A\shortto B}y$. Then, since $y\in \mods{A\wedge \neg B}$,  by Success, $A\rightarrow B\in\bel{(\Psi\ast A\shortto B)\ast C}$. Hence, by \pprincRev{3}, $A\rightarrow B\in\bel{(\Psi\ast A\shortTo B)\ast C}$. Once again, by Success and the fact that $y\in \mods{A\wedge \neg B}$, $x \prec_{\Psi\ast A\shortTo B}y$, as required.

\end{itemize}
\item[{\bf (c)}] \pprincRevR{4} and \pprincRev{4}: Similar to (c) above.

\end{itemize}
\end{myproof}

\vspace{1em}

%=====================================================

\NoAddedFactualBeliefs*

\begin{myproof}
We need to show that that  $\min(\preccurlyeq_{\Psi\ast A\shortTo B}, W) = \min(\preccurlyeq_{\Psi\ast A\shortto B}, W)$. 

We first prove $\min(\preccurlyeq_{\Psi\ast A\shortTo B}, W) \subseteq \min(\preccurlyeq_{\Psi\ast A\shortto B}, W)$. Assume that $y\in  \min(\preccurlyeq_{\Psi\ast A\shortTo B}, W)$ but, for reductio, that $y\notin  \min(\preccurlyeq_{\Psi\ast A\shortto B}, W)$. From the latter and \SRevR, there exists $x\in \mods{A\shortto B}$ such that $x\prec_{\Psi\ast A\shortto B}y$. If $y\in \mods{A\shortto B}$, then $x \prec_{\Psi\ast A\shortTo B} y$ by \pprincRevR{1}. If $y\in \mods{A\wedge \neg B}$, then $x \prec_{\Psi\ast A\shortTo B} y$ by \pprincRevR{3}. Either way, $y\notin  \min(\preccurlyeq_{\Psi\ast A\shortTo B}, W)$. Contradiction. Hence $y\in  \min(\preccurlyeq_{\Psi\ast A\shortto B}, W)$ and so  $\min(\preccurlyeq_{\Psi\ast A\shortTo B}, W) \subseteq \min(\preccurlyeq_{\Psi\ast A\shortto B}, W)$, as required.

We now show that $\min(\preccurlyeq_{\Psi\ast A\shortto B}, W) \subseteq \min(\preccurlyeq_{\Psi\ast A\shortTo B}, W)$. Assume that $y\in  \min(\preccurlyeq_{\Psi\ast A\shortto B}, W)$ but, for reductio, that $y\notin  \min(\preccurlyeq_{\Psi\ast A\shortTo B}, W)$. From the former and \SRevR, we have  $y\in \mods{A\shortto B}$. From the latter, there exists $x\in W$ such that $x \prec_{\Psi\ast A\shortTo B} y$.  If $x\in \mods{A\shortto B}$, then $x \prec_{\Psi\ast A\shortto B} y$ by \pprincRevR{1}.  If $x\in \mods{A\wedge \neg B}$, then $x \prec_{\Psi\ast A\shortto B} y$ by \pprincRevR{4}. Either way, $y\notin  \min(\preccurlyeq_{\Psi\ast A\shortto B}, W)$. Contradiction. Hence $y\in  \min(\preccurlyeq_{\Psi\ast A\shortTo B}, W)$ and so  $\min(\preccurlyeq_{\Psi\ast A\shortto B}, W) \subseteq \min(\preccurlyeq_{\Psi\ast A\shortTo B}, W)$, as required.
\end{myproof}

\vspace{1em}

%=====================================================

\PropertiesThree*

\begin{myproof}
For this proof, we will make use of the fact that elementary operators satisfy the following principle from \cite{DBLP:conf/kr/0001C18}: 

\begin{tabbing}
\=BLAHI: \=\kill

\> \BetaRevR{2}  \> If $x \not\in \min(\preccurlyeq_{\Psi}, \mods{C})$, $x \in \mods{A}$, $y\in \mods{\neg A}$, and\\
\>\> $y \prec_{\Psi \ast A} x$, then $y \prec_{\Psi\ast C} x$\\[-0.25em]
\end{tabbing} 
\vspace{-0.75em}

\noindent We will also make use of the concept of a {\em state restriction}. Informally, the restriction $\Psi\,\mid\, A$ of a state $\Psi$ to a sentence $A$ corresponds to the agent's categorically ruling out as impossible any world inconsistent with $A$. In terms of the TPO associated with a restricted state, we define $\preccurlyeq_{\Psi\mid A}$ as $\preccurlyeq_{\Psi}\cap (\mods{A}\times \mods{A})$. This notion allows us to articulate the following property, which is also satisfied by elementary operators:

\begin{tabbing}
\=BLAHI: \=\kill

\> (SR)  \>  (1) If  $x\in\min(\preccurlyeq_{\Psi\mid A}, \mods{B})$ and $y\in\mods{\neg B}$, then\\
\>\> $x\prec_{(\Psi\mid A) \ast B}y$  \\[0.1cm]
\> \> (2) Otherwise $x\preccurlyeq_{(\Psi\mid A) \ast B}y$ iff $x \preccurlyeq_{\Psi\ast B}y$\\[-0.25em]
\end{tabbing} 
\vspace{-0.75em}

\noindent Assume that $x, y\in\mods{A}$. We want to show that $x\preccurlyeq_{\Psi\ast A\shortTo B}y$ iff $x\preccurlyeq_{(\Psi\mid A) \ast B}y$.

\begin{itemize}

\item[{\bf (a)}] $x, y\in\mods{B}$: Then $x, y \in\mods{A\shortto B}$, so, by \pprincRevR{1}, $x\preccurlyeq_{\Psi\ast A\shortTo B}y$ iff $x\preccurlyeq_{\Psi\ast A\shortto B}y$ and, by \CRevR{1}, $x\preccurlyeq_{\Psi\ast A\shortto B}y$ iff $x\preccurlyeq_{\Psi} y$. Since $x, y\in\mods{A}$ and $x, y\in\mods{B}$, by \CRevR{1} and (SR), $x\preccurlyeq_{\Psi} y$ iff $x\preccurlyeq_{(\Psi\mid A) \ast B}y$. 

\item[{\bf (b)}] $x, y\in\mods{\neg B}$: Similar to (a), using \pprincRevR{2} and \CRevR{2}.

\item[{\bf (c)}] $x\in\mods{B}$, $y\in\mods{\neg B}$: If $x\in\min(\preccurlyeq_{\Psi\mid A}, \mods{B})$, then, by (SR), we have $x\prec_{(\Psi\mid A) \ast B} y$. Furthermore, since $x\in\mods{A}$, it follows from $x\in\min(\preccurlyeq_{\Psi\mid A}, \mods{B})$ that  $x\in\min(\preccurlyeq_{\Psi}, \mods{A\wedge B})$. By Success and the fact that $y\in\mods{A\wedge\neg B}$, we then recover $x\prec_{\Psi\ast A\shortTo B}y$. Hence $x\preccurlyeq_{\Psi\ast A\shortTo B}y$ iff $x\preccurlyeq_{(\Psi\mid A) \ast B}y$, as required.

So assume that $x\notin\min(\preccurlyeq_{\Psi\mid A}, \mods{B})$, then (SR) gives us $x\preccurlyeq_{(\Psi\mid A) \ast B}y$ iff $x\preccurlyeq_{\Psi\ast B}y$. We now show that $x \preccurlyeq_{\Psi\ast A\shortto B} y$ iff $x \preccurlyeq_{\Psi\ast B}y$.

We know from $x\in\mods{A}$ that $x\in\min(\preccurlyeq,\mods{A\shortto B})$ iff $x\in\min(\preccurlyeq,\mods{B})$. So we consider two cases:
\begin{itemize}

\item[{\bf (i)}] $x\in\min(\preccurlyeq_{\Psi},\mods{A\shortto B})$ and $x\in\min(\preccurlyeq_{\Psi},\mods{B})$: Since $y\in\mods{A\wedge\neg B}$, by Success, we have $x \prec_{\Psi\ast A\shortto B} y$ and $x \prec_{\Psi\ast B}y$. Hence  $x \preccurlyeq_{\Psi\ast A\shortto B} y$ iff $x \preccurlyeq_{\Psi\ast B}y$.
 
\item[{\bf (ii)}]   $x\notin\min(\preccurlyeq_{\Psi},\mods{A\shortto B})$ and $x\notin\min(\preccurlyeq_{\Psi},\mods{B})$: We'll establish our biconditional by showing that $y \prec_{\Psi\ast A\shortto B} x$ iff $y \prec_{\Psi\ast B}x$. We recover the implication from $y \prec_{\Psi\ast A\shortto B} x$ to $y \prec_{\Psi\ast B}x$, by \BetaRevR{2}, since $x\notin\min(\preccurlyeq_{\Psi},\mods{B})$, $x\in\mods{A\shortto B}$ and $y\in\mods{A\wedge\neg B}$. We recover the converse implication by the same principle, this time because  $x\notin\min(\preccurlyeq_{\Psi},\mods{A\shortto B})$, $x\in\mods{ B}$ and $y\in\mods{\neg B}$.
%We additionally know, from $y\in\mods{\neg B}$ that $y\notin\min(\preccurlyeq,\mods{A\shortto B})$ and $y\notin\min(\preccurlyeq,\mods{B})$. 

\end{itemize}
So we have established that $x \preccurlyeq_{\Psi\ast A\shortto B} y$ iff $x \preccurlyeq_{\Psi\ast B}y$. Since we already have $x\preccurlyeq_{(\Psi\mid A) \ast B}y$ iff $x\preccurlyeq_{\Psi\ast B}y$, it then follows that $x\preccurlyeq_{(\Psi\mid A) \ast B}y$ iff $x \preccurlyeq_{\Psi\ast A\shortto B} y$.

We now note the following consequence of Definition \ref{MyProposal}:
\begin{itemize}
\item[~] If $x, y\in\mods{A}$, then 
\begin{itemize}

\item[(1')] If $x\in\min(\preccurlyeq_{\Psi}, \mods{A\wedge B})$ and $y\in\mods{A\wedge \neg B}$, then $x\prec_{\Psi\ast A\shortTo B}y$

\item[(2')] Otherwise $x\preccurlyeq_{\Psi\ast A\shortTo B}y$ iff $x\preccurlyeq_{\Psi\ast A\shortto B} y$

\end{itemize}
\end{itemize}

So, if $x\notin\min(\preccurlyeq_{\Psi}, \mods{A\wedge B})$, it follows from $x\in\mods{B}$ and $y\in\mods{\neg B}$, by (1'), that $x\preccurlyeq_{\Psi\ast A\shortTo B}y$ iff $x\preccurlyeq_{\Psi\ast A\shortto B} y$ and so $x\preccurlyeq_{\Psi\ast A\shortTo B}y$ iff $x\preccurlyeq_{(\Psi\mid A) \ast B}y$, as required.

If $x\in\min(\preccurlyeq_{\Psi}, \mods{A\wedge B})$, then $x\prec_{\Psi\ast A\shortTo B}y$, by (2'). But we will also have $x\in\min(\preccurlyeq_{\Psi\mid A}, \mods{B})$ and so, by (SR), $x\prec_{(\Psi\mid A) \ast B}y$. Hence $x\preccurlyeq_{\Psi\ast A\shortTo B}y$ iff $x\preccurlyeq_{(\Psi\mid A) \ast B}y$, as required. 
%Since, as we have established, $x \preccurlyeq_{\Psi\ast A\shortto B} y$ iff $x \preccurlyeq_{\Psi\ast B}y$, this gives us:
% \begin{itemize}
%
%\item[(1'')] If $x\in\mods{A\wedge \neg B}$, $y\in\mods{A\wedge B}$, and $x\preccurlyeq_{\Psi\ast B}y$, then  $y\prec_{\Psi\ast A\shortTo B}x$
%
%\item[(2'')] Otherwise $x\preccurlyeq_{\Psi\ast A\shortTo B}y$ iff $x\preccurlyeq_{\Psi\ast B}y$
%
%\end{itemize}

\item[{\bf (d)}] $x\in\mods{\neg B}$, $y\in\mods{B}$: Similar to (c).

\end{itemize}
\end{myproof}

\vspace{1em}

%=====================================================

\NatAndRatClosure*

\begin{myproof}
%We adapt and expand the proof of Theorem 4 of \jake{Chandler \& Booth (IJCAI2019 submission)}, which essentially establishes a very special case of this.
We want to show that our proposal has the result that $\preccurlyeq_{\Psi\astN A\shortTo B}$ is the `flattest' TPO--in a technical sense to be defined below--such that the following lower bound constraint is satisfied, given $A\shortTo \neg B\notin\cbel{\Psi}$:

\begin{tabbing}
\=BLAHBLI: \=\kill
\> \> $ \cbel{\Psi} \cup \{A\shortTo B\}\subseteq \cbel{\Psi \astN A\shortTo B}$ \\[-0.25em]
\end{tabbing} 
\vspace{-0.75em}

\noindent In view of Definitions 20 and 21 of \cite{lehmann1992does}, the upshot of this is then that $\preccurlyeq_{\Psi\astN A\shortTo B}$ is the unique TPO corresponding to the rational closure of $\cbel{\Psi}\cup\{A\shortTo B\}$. 

We first note that, given AGM, the syntactic condition $A\shortTo \neg B\notin\cbel{\Psi}$ corresponds to the semantic condition $\min(\preccurlyeq_{\Psi}, \mods{A})\nsubseteq\mods{\neg B}$, while  the lower bound condition can be expressed as follows:

\begin{tabbing}
\=BLAHBLI: \=\kill
\> \> (a) If $x \prec_{\Psi} y$, then $x \prec_{\Psi \astN A\shortTo B} y$ and \\[0.1cm]
\> \> (b) $\min(\preccurlyeq_{\Psi\astN A\shortTo B},\mods{A})\subseteq \mods{B}$ \\[-0.25em]
\end{tabbing} 
\vspace{-0.75em}

\noindent Indeed, the lower bound constraint simply amounts to the conjunction of Success, which is equivalent to (b), with the claim that $ \cbel{\Psi} \subseteq \cbel{\Psi \astN A\shortTo B}$, which is equivalent to (a). 
With this in hand, we now prove two lemmas. First: 

\begin{restatable}{lem}{SPP}
\label{SPP}
When the extension of $\astN$ to $L_c$ is defined as in Definition \ref{MyProposal}, if $\min(\preccurlyeq_{\Psi}, \mods{A})\nsubseteq\mods{\neg B}$,  then $\preccurlyeq_{\Psi \astN A\shortto B}$ satisfies the lower bound condition.
\end{restatable}

\noindent We have already established (unconditional) satisfaction of (b) in Theorem  \ref{Representation}. So we just need to establish satisfaction of (a). 

Assume $\min(\preccurlyeq_{\Psi}, \mods{A})\nsubseteq\mods{\neg B}$ and  $x\prec_{\Psi}y$. We need to establish that  $x \prec_{\Psi \astN A\shortTo B} y$. To do this, we will first establish that $x~\prec_{\Psi \astN A\shortto B}~y$. Given the definition of $\astN$, we will have $x~\prec_{\Psi \astN A\shortto B}~y$ iff either

\begin{itemize}

\item[(1)] $x \in \min(\preccurlyeq_{\Psi}, \mods{A\shortto B})$ and $y \notin \min(\preccurlyeq_{\Psi}, \mods{A\shortto B})$, or 

\item[(2)] $x, y \notin \min(\preccurlyeq_{\Psi}, \mods{A\shortto B})$ and $x\prec_{\Psi}y$

\end{itemize}

\noindent Assume for reductio that $y \in \min(\preccurlyeq_{\Psi}, \mods{A\shortto B})$. Then, since $x\prec_{\Psi}y$, we must have  $\min(\preccurlyeq_{\Psi}, W)\subseteq \mods{A\wedge\neg B}$. But we have assumed $\min(\preccurlyeq_{\Psi}, \mods{A})\nsubseteq\mods{\neg B}$. Contradiction. Hence $y \notin \min(\preccurlyeq_{\Psi}, \mods{A\shortto B})$.  This leaves us with two possibilities. The first is that $x, y \notin \min(\preccurlyeq_{\Psi }, \mods{A\shortto B})$, which, given $x\prec_{\Psi}y$,  places us in case (2). The second is that $x \in \min(\preccurlyeq_{\Psi}, \mods{A\shortto B})$ and $y \notin \min(\preccurlyeq_{\Psi}, \mods{A\shortto B})$, which places us in case (1). Either way, then, $x~\prec_{\Psi \astN A\shortto B}~y$, as required. 

We now need to show, from this, that, given Definition \ref{MyProposal}, it follows that  $x~\prec_{\Psi \astN A\shortTo B}~y$. If (i) $x, y\in\mods{A\shortto B}$, (ii) $x, y\in\mods{A\wedge\neg B}$ or (iii) $x\in\mods{A\shortto B}$ and $y\in\mods{A\wedge\neg B}$, then the required result follows from \pprincRevR{1}, \pprincRevR{2} and \pprincRevR{3}, respectively. So assume $x\in\mods{A\wedge\neg B}$ and $y\in\mods{A\shortto B}$. We will now show that $y\notin D(\preccurlyeq_{\Psi\ast A\shortto B}, A\wedge B)$. Given this, the required conclusion will follow by \pprincRevR{5}. So assume for reductio that $y\in D(\preccurlyeq_{\Psi\ast A\shortto B}, A\wedge B)$, i.e.~that $y\preccurlyeq_{\Psi\astN A\shortto B} z$, for all $z\in \min(\preccurlyeq_{\Psi\astN A\shortto B}, \mods{A\wedge B})$. Since $x~\prec_{\Psi \astN A\shortto B}~y$, we will therefore have $x~\prec_{\Psi \astN A\shortto B}~z$, for all $z\in \min(\preccurlyeq_{\Psi\astN A\shortto B}, \mods{A\wedge B})$.  Since $x\in\mods{A\wedge \neg B}$ and $z\in\mods{A\shortto B}$, by \CRevR{4}, we recover $x\prec_{\Psi} z$, for all $z\in \min(\preccurlyeq_{\Psi\astN A\shortto B}, \mods{A\wedge B})$.  However, by \CRevR{1}, we have the result that $\min(\preccurlyeq_{\Psi}, \mods{A\wedge B})= \min(\preccurlyeq_{\Psi\astN A\shortto B}, \mods{A\wedge B})$. So $x\prec_{\Psi} z$ for all $z\in \min(\preccurlyeq_{\Psi}, \mods{A\wedge B})$. But, since  $x\in\mods{A\wedge\neg B}$, this contradicts our assumption that $\min(\preccurlyeq_{\Psi}, \mods{A})\nsubseteq\mods{\neg B}$. Hence $y\notin D(\preccurlyeq_{\Psi\ast A\shortto B}, A\wedge B)$, as required. This completes the proof of Lemma \ref{SPP}.

For our second lemma, we will make use of the convenient representation of TPOs by their corresponding {\em ordered partitions} of $W$. The ordered partition $\langle S_1, S_2, \ldots S_m\rangle$ of $W$ corresponding to a TPO $\preccurlyeq$ is such that $x \preccurlyeq y$ iff $r(x, \preccurlyeq) \leq$ \mbox{$r(y, \preccurlyeq)$,} where $r(x, \preccurlyeq)$ denotes the `rank' of $x$ with respect to $\preccurlyeq$ and is defined by taking $S_{r(x, \preccurlyeq)}$ to be the cell in the partition that contains $x$. 

This lemma is given as follows:

\begin{restatable}{lem}{FlattestLI}
\label{FlattestLI}
If $\min(\preccurlyeq_{\Psi}, \mods{A})\nsubseteq\mods{\neg B}$, then $\preccurlyeq_{\Psi \astN A\shortTo B}~\sqsupseteq~~\preccurlyeq$, for any TPO $\preccurlyeq$ satisfying the lower bound condition. 
\end{restatable}

\noindent where:

\begin{definition}
\label{dfn:Flatter}
$\sqsupseteq$ is a binary relation on the set of TPOs over $W$ such such that, for any TPOs $\preccurlyeq_1$ and $\preccurlyeq_2$, whose corresponding ordered partitions are given by $\langle S_1, S_2, \ldots, S_m \rangle$ and $\langle T_1, T_2, \ldots, T_m \rangle$ respectively,   
$
\preccurlyeq_1~
\sqsupseteq~
 \preccurlyeq_2
$
iff
either (i) $S_i = T_i$ for all $i = 1, \ldots, m$, or (ii) $S_i \supset T_i$ for the first $i$ such that $S_i \neq T_i$. 
\end{definition}

\noindent $\sqsupseteq$ partially orders the set of TPOs over $W$ according to  comparative `flatness', with the flatter TPOs appearing `greater' in the ordering, so that $\preccurlyeq_1~\sqsupseteq~\preccurlyeq_2$ iff $\preccurlyeq_1 $ is at least as as flat as $\preccurlyeq_2$.

Let $\langle T_1,\ldots, T_m\rangle$ be the ordered partition corresponding to the TPO $\preccurlyeq_{\Psi \astN A\shortTo B}$. Let $\preccurlyeq $ be any TPO satisfying the lower bound condition: 

\begin{tabbing}
\=BLAHBLI: \=\kill
\> \> (a) If $x \prec_{\Psi} y$, then $x \prec y$ and \\[0.1cm]
\> \> (b) $\min(\preccurlyeq,\mods{A})\subseteq \mods{B}$ \\[-0.25em]
\end{tabbing} 
\vspace{-0.75em}

\noindent Let $\langle S_1,\ldots, S_n\rangle$ be its corresponding ordered partition. We must show that the following relation holds:  $\preccurlyeq_{\Psi \astN A\shortTo B}~\sqsupseteq~\preccurlyeq$.

If $T_i=S_i$ for all $i$, then we are done. So let $i$ be minimal such that $T_i\neq S_i$. We must show $S_i\subset T_i$. So let $y\in S_i$ and assume, for contradiction, that  $y\notin T_i$. We know that $T_i\neq \varnothing$, since, otherwise, $\bigcup_{j < i} T_j = W$, hence $\bigcup_{j < i} S_j = W$ and so $S_i=\varnothing$, contradicting $S_i\neq T_i$. So let $x\in T_i$. Then, since $y\notin T_i$, we have $x \prec_{\Psi \astN A\shortTo B} y$. We are going to show that this entails that $\exists z$ such that 
\begin{itemize}

\item[(i)] $z\sim_{\Psi \astN A\shortTo B} x$, i.e.~$z\in T_i$, and 

\item[(ii)]  $z\prec y$. 

\end{itemize}
But if it were the case that $z\prec y$, then, since $y\in S_i$, $z\in \bigcup_{j < i} S_j = \bigcup_{j < i} T_j$, contradicting $z\in T_i$. Hence $y\in T_i$ and so we can conclude that $S_i\subset T_i$, as required.

So assume $\min(\preccurlyeq_{\Psi}, \mods{A})\nsubseteq\mods{\neg B}$. From  $x \prec_{\Psi \astN A\shortTo B} y$ we know that both 
\begin{itemize}

\item[(1)]   (a) $y\notin D(\preccurlyeq_{\Psi\ast A\shortto B}, A\wedge B)$ or 

(b) $y\in\mods{A\wedge \neg B}$ or 

 (c) $x\in D(\preccurlyeq_{\Psi\ast A\shortto B}, A\wedge B)\cap\mods{A\shortto B}$, 
 
\end{itemize}
and 
\begin{itemize}

\item[(2)]   (a) $y\in D(\preccurlyeq_{\Psi\ast A\shortto B}, A\wedge B)\cap\mods{A\shortto B}$ iff $x\notin D(\preccurlyeq_{\Psi\ast A\shortto B}, A\wedge B)\cap\mods{A\shortto B}$ or 

(b) $x \prec_{\Psi\ast A\shortto B} y$
 
\end{itemize}
Assume (2) (b), i.e.~$x \prec_{\Psi\ast A\shortto B} y$. By the definition of $\astN$, either:
\begin{itemize}

\item[(3)] $x \in \min(\preccurlyeq_{\Psi}, \mods{A\shortto B})$ and $y \notin \min(\preccurlyeq_{\Psi}, \mods{A\shortto B})$, or 

\item[(4)] $x, y \notin \min(\preccurlyeq_{\Psi}, \mods{A\shortto B})$ and $x\prec_{\Psi}y$

\end{itemize}
 If (4), then from $x\prec_{\Psi} y$ and (a) of the lower bound condition, it follows that $x\prec y$ and we are done, with $x$ playing the role of $z$ in (i) and (ii) above. So assume (3). We now split into two cases:
  \begin{itemize}

\item[-] Assume $y\in\mods{A\shortto B}$. Then from $x \in  \mods{A\shortto B}$ (which follows from (3)), \CRevR{1} and $x \prec_{\Psi\ast A\shortto B} y$,  it follows that $x\prec_{\Psi} y$. From this and (a) of the lower bound condition, we then again have $x\prec y$ and we are done, with $x$ playing the role of $z$ in (i) and (ii) above. 

\item[-] Assume $y\in\mods{A\wedge \neg B}$. If $x\in\min(\preccurlyeq, W)$, then, since, by (b) of the lower bound condition, $\min(\preccurlyeq,\mods{A})\subseteq \mods{B}$, we have $x \prec y$ and we are done, with $x$ playing the role of $z$ in (i) and (ii) above.  So assume  $x\notin\min(\preccurlyeq, W)$. Let $z\in\min(\preccurlyeq, W)$. Then $z\preccurlyeq x$. By the contrapositive of (a) of the lower bound condition, $z\preccurlyeq_{\Psi} x$. If $z\in\mods{A}$, then,  by $\min(\preccurlyeq, \mods{A})\subseteq \mods{B}$, we have $z\in\mods{A\shortto B}$. Obviously, if $z\in\mods{\neg A}$, then again $z\in\mods{A\shortto B}$. So  $z\in\mods{A\shortto B}$. Since, furthermore,  $x \in \min(\preccurlyeq_{\Psi}, \mods{A\shortto B})$ (by (3)) and $z\preccurlyeq_{\Psi} x$, it follows from this that $z \in \min(\preccurlyeq_{\Psi}, \mods{A\shortto B})$. By \CRevR{1}, we then have $z\sim_{\Psi\astN A\shortto B} x$. From $\min(\preccurlyeq, \mods{A})\subseteq \mods{B}$ and $y\in\mods{A\wedge \neg B}$, $y\notin\min(\preccurlyeq, W)$. Hence, since  $z\in\min(\preccurlyeq, W)$, it follows that $z\prec  y$. So $z$ satisfies conditions (i) and (ii) above and we are done.
 
  \end{itemize}
 Assume (2)(a), i.e.~$y\in D(\preccurlyeq_{\Psi\ast A\shortto B}, A\wedge B)\cap\mods{A\shortto B}$ iff $x\notin D(\preccurlyeq_{\Psi\ast A\shortto B}, A\wedge B)\cap\mods{A\shortto B}$. We now split into three cases:
\begin{itemize}

\item[-]  Assume (1)(a), i.e.~$y\notin D(\preccurlyeq_{\Psi\ast A\shortto B}, A\wedge B)$. Then, by (2)(a), $x\in D(\preccurlyeq_{\Psi\ast A\shortto B}, A\wedge B)$ and $x\in\mods{A\shortto B}$. From the first two facts, we have $x\prec_{\Psi\ast A\shortto B}y$. Since we can see from the item immediately below that, if $y\in\mods{A\wedge \neg B}$, then we recover the required result, we can assume $y\in\mods{A\shortto B}$. From this, the fact that $x\in\mods{A\shortto B}$, and $x\prec_{\Psi\ast A\shortto B}y$, we recover $x\prec_{\Psi}y$ by \CRevR{1}. From this, by (a) of the lower bound principle, we obtain $x\prec y$ and we are done, with $x$ playing the role of $z$ in (i) and (ii) above. 

\item[-] Assume (1)(b), i.e.~$y\in\mods{A\wedge \neg B}$. If $x\prec y$, then we are done, with $x$ playing the role of $z$ in (i) and (ii) above.  So assume $y\preccurlyeq x$, from which it follows by the contrapositive of (a) of the  lower bound condition  that $y\preccurlyeq_{\Psi} x$. By the definition of $\astN$, we then have $y\preccurlyeq_{\Psi\astN A\shortto B} x$ iff $x\notin\min(\preccurlyeq_{\Psi},\mods{A\shortto B})$. Now, as we show above, if $x\in\min(\preccurlyeq_{\Psi},\mods{A\shortto B})$ and $y\in\mods{A\wedge \neg B}$, then we are done. So assume $x\notin\min(\preccurlyeq_{\Psi},\mods{A\shortto B})$ and hence $y\preccurlyeq_{\Psi\astN A\shortto B} x$. Let $z\in\min(\preccurlyeq_{\Psi}, \mods{A})$. Since $\min(\preccurlyeq_{\Psi}, \mods{A})\nsubseteq\mods{\neg B}$ and $y\in\mods{A\wedge \neg B}$, it follows that $z\preccurlyeq_{\Psi} y$. By \CRevR{4}, since $x\in \mods{A\shortto B}$ and $y\in\mods{A\wedge \neg B}$, we then have $z\preccurlyeq_{\Psi\astN A\shortto B} y$. Since, furthermore, $y\preccurlyeq_{\Psi\astN A\shortto B} x$, we then have $z\preccurlyeq_{\Psi\astN A\shortto B} x$. But from $x\in D(\preccurlyeq_{\Psi\ast A\shortto B}, A\wedge B)$, it follows, by definition, that $x\preccurlyeq_{\Psi\astN A\shortto B} z$ and hence $z\sim_{\Psi\astN A\shortto B} x$. Therefore $z$ satisfies condition (i) above.

 It now remains to be shown that $z$ satisfies condition (ii) above, i.e.~that $z\prec y$. We first show that it follows from (a) of the lower bound condition that $\min(\preccurlyeq_{\Psi}, \mods{A})\cap \min(\preccurlyeq, \mods{A})\neq \varnothing$. Indeed, we can show that, given (a), if $x\in \min(\preccurlyeq_{\Psi}, \mods{A})$ but $x\notin \min(\preccurlyeq, \mods{A})$, then there exists a distinct $z\in \min(\preccurlyeq_{\Psi}, \mods{A})$, such that $z\prec x$.  Since we have assumed that $W$ is finite and hence so is $\min(\preccurlyeq_{\Psi}, \mods{A})$, it cannot be the case that such a $z$ exists for all $x\in \min(\preccurlyeq_{\Psi}, \mods{A})$. Hence $\min(\preccurlyeq_{\Psi}, \mods{A})\cap \min(\preccurlyeq, \mods{A})\neq \varnothing$. 
%  So assume then that $x\in \min(\preccurlyeq_{\Psi}, \mods{A})$ and $x\notin \min(\preccurlyeq, \mods{A})$.  Let $z\in \min(\preccurlyeq, \mods{A})$, so that $z\prec x$ (and hence $z\neq x$). Assume for reductio that $z\notin \min(\preccurlyeq_{\Psi}, \mods{A})$. Then, since $z\in \mods{A}$ but  $x\in \min(\preccurlyeq_{\Psi}, \mods{A})$, $x\prec_{\Psi} z$, and hence, by (a) of the lower bound condition, $x\prec z$. Since $x\in \mods{A}$, this contradicts $z\in \min(\preccurlyeq, \mods{A})$. Hence $z\in \min(\preccurlyeq_{\Psi}, \mods{A})$, as required. \jake{no idea what went on here: must be a copy paste issue or something}
In view of this, we are free to assume that $z\in\min(\preccurlyeq, \mods{A})$. Since $\min(\preccurlyeq,\mods{A})\subseteq \mods{B}$ and $y\in\mods{A\wedge \neg B}$, we then have $z\prec y$, as required.

\item[-] Assume (1)(c), i.e.~$x\in D(\preccurlyeq_{\Psi\ast A\shortto B}, A\wedge B)\cap\mods{A\shortto B}$. Then, by (2)(a), we are in one of the two cases immediately above and the required result follows.
 
\end{itemize}
This completes the proof of Lemma \ref{FlattestLI} and hence the proof of Proposition \ref{NatAndRatClosure}.
\end{myproof}

%=====================================================

\closestNat*

\begin{myproof}
Note first that the distance between TPOs over $W$ is simply the sum of the distances between the restrictions of these TPOs to the different pairs drawn from $W$.

We assume $\preccurlyeq_{\Psi\astN A}\neq \preccurlyeq_{\Psi\star A}$. We have three cases to consider:
\begin{itemize}

\item[-] $x, y\in \min(\preccurlyeq_{\Psi}, \mods{A})$: Since both $\astN$ and $\star$ satisfy AGM, we have $\min(\preccurlyeq_{\Psi\star A}, W) = \min(\preccurlyeq_{\Psi}, \mods{A}) = \min(\preccurlyeq_{\Psi\astN A}, W) $. Hence $d_K(\preccurlyeq_{\Psi\astN A}\cap \{x, y\}, \preccurlyeq_{\Psi}\cap \{x, y\}) = d_K(\preccurlyeq_{\Psi\star A}\cap \{x, y\}, \preccurlyeq_{\Psi}\cap \{x, y\})$.

\item[-] $x, y\notin \min(\preccurlyeq_{\Psi}, \mods{A})$: We know that, by definition, for all $x, y \notin \min(\preccurlyeq_{\Psi}, \mods{A})$, $x \preccurlyeq_{\Psi \astN A} y$ iff $x \preccurlyeq_{\Psi} y$, so that $\preccurlyeq_{\Psi\astN A}\cap \{x, y\} = \preccurlyeq_{\Psi}\cap \{x, y\}$. Hence $d_K(\preccurlyeq_{\Psi\astN A}\cap \{x, y\}, \preccurlyeq_{\Psi}\cap \{x, y\}) = 0$.  Since $\preccurlyeq_{\Psi\astN A}\neq \preccurlyeq_{\Psi\star A}$, but $\min(\preccurlyeq_{\Psi\star A}, W) = \min(\preccurlyeq_{\Psi\astN A}, W) $, we have $\preccurlyeq_{\Psi\star A}\cap  \{x, y\} \neq \preccurlyeq_{\Psi\astN A}\cap  \{x, y\} = \preccurlyeq_{\Psi}\cap  \{x, y\}$. So $d_K(\preccurlyeq_{\Psi\star A}\cap  \{x, y\}, \preccurlyeq_{\Psi}\cap  \{x, y\}) > 0$ and hence $d_K(\preccurlyeq_{\Psi\astN A}\cap \{x, y\}, \preccurlyeq_{\Psi}\cap \{x, y\}) < d_K(\preccurlyeq_{\Psi\star A}\cap \{x, y\}, \preccurlyeq_{\Psi}\cap \{x, y\})$.

\item[-] $x\in \min(\preccurlyeq_{\Psi}, \mods{A})$ and $y\notin \min(\preccurlyeq_{\Psi}, \mods{A})$:  Since both $\astN$ and $\star$ satisfy AGM, we have $\min(\preccurlyeq_{\Psi\star A}, W) = \min(\preccurlyeq_{\Psi}, \mods{A}) = \min(\preccurlyeq_{\Psi\astN A}, W) $. So $x \prec_{\Psi\astN A} y$ and $x \prec_{\Psi\star A} y$ and hence $d_K(\preccurlyeq_{\Psi\astN A}\cap \{x, y\}, \preccurlyeq_{\Psi}\cap \{x, y\}) = d_K(\preccurlyeq_{\Psi\star A}\cap \{x, y\}, \preccurlyeq_{\Psi}\cap \{x, y\})$.

\end{itemize}
From the above, it then follows that  $d_K(\preccurlyeq_{\Psi\astN A}, \preccurlyeq_{\Psi})< d_K(\preccurlyeq_{\Psi\star A}, \preccurlyeq_{\Psi})$, as required.
\end{myproof}

\vspace{1em}

%=====================================================

\HanssonProblem*

\begin{myproof}
We first recall the fact that \KRev{8} is equivalent, given the remainder of the AGM postulates, to the principle of Disjunctive Inclusion:
\begin{tabbing}
\=BLAHHI: \=\kill

\> $\mathrm{(DI^\ast)}$ \> If $\neg A\notin \bel{\Psi\ast A\vee C }$, then $\bel{\Psi\ast A\vee C }\subseteq$\\
\>\>$\bel{\Psi\ast A}$ \\[-0.25em]

\end{tabbing} 
\vspace{-1em}

\noindent Consider now the initial TPO $\preccurlyeq_{\Psi}$ depicted in Figure \ref{fig:HanssonInit}.

\begin{figure}[!h]
\begin{centering}
\begin{tikzpicture}[->,>=stealth']
\scalebox{1}{%
 \node[state,
  text width=2cm,        % max text width
 ] (astN)   
 {
 \scalebox{0.6}{
\begin{tabular}[c]{c|c|c}
     \multicolumn{1}{c|}{$A\wedge B$} &  $\neg A$   &   \multicolumn{1}{c}{$A\wedge \neg B$} \\
    \hline
    \rule{0pt}{2em}
      2   &   &  \\
         & 4  &  \\
         &   & 3 \\
         &  1  &  \\[0.25em]
  \end{tabular}
}
};

}
\end{tikzpicture}
\caption{Initial TPO $\preccurlyeq_{\Psi}$ , in relation to proof of Proposition \ref{HanssonProblem}}
\label{fig:HanssonInit}
\end{centering}
\end{figure}

\noindent It is easily verified that the two closest TPOs to this one, subject to the constraint that $A\shortTo B$ is in the associated conditional belief set, are given by, the TPOs $\preccurlyeq_{\Theta 1}$, pictured in Figure \ref{fig:HanssonClosest} on the left, and $\preccurlyeq_{\Theta 2}$, pictured  to its right:

\begin{figure}[!h]
\begin{centering}
\begin{tikzpicture}[->,>=stealth']
\scalebox{1}{%
 \node[state,
  text width=2cm,        % max text width
 ] (astN)   
 {
 \scalebox{0.6}{
\begin{tabular}[c]{c|c|c}
     \multicolumn{1}{c|}{$A\wedge B$} &  $\neg A$   &   \multicolumn{1}{c}{$A\wedge \neg B$} \\
    \hline
    \rule{0pt}{2em}
             &   & 3 \\
      2   &   &  \\
         & 4  &  \\
         &  1  &  \\[0.25em]
  \end{tabular}
}
};
 \node[state,
  text width=2cm,        % max text width
  right of = astN,
   node distance=3cm,     % distance to K
 ] (blah)   
 {
 \scalebox{0.6}{
\begin{tabular}[c]{c|c|c}
     \multicolumn{1}{c|}{$A\wedge B$} &  $\neg A$   &   \multicolumn{1}{c}{$A\wedge \neg B$} \\
    \hline
    \rule{0pt}{2em}
         & 4  &  \\
         &   & 3 \\
      2   &   &  \\
         &  1  &  \\[0.25em]
  \end{tabular}
}
};

}
\end{tikzpicture}
\caption{Closest TPOs to initial TPO $\preccurlyeq_{\Psi}$, in relation to proof of Proposition \ref{HanssonProblem}}
\label{fig:HanssonClosest}
\end{centering}
\end{figure}

\noindent Let $\mods{X} = \{2, 3, 4\}$ and $\mods{Y} = \{3, 4\}$. From the diagram, we can see that the following hold: 
\begin{itemize}

\item[] (i) $(X\vee Y)\shortTo \neg Y\notin \cbel{\Theta_1}\cap \cbel{\Theta_2}$

\item[] (ii) $(X\vee Y)\shortTo \neg 3\in \cbel{\Theta_1}\cap \cbel{\Theta_2}$

\item[] (iii) $Y\shortTo \neg 3\notin \cbel{\Theta_1}\cap \cbel{\Theta_2}$

\end{itemize}

\noindent But this contradicts $\mathrm{(DI^\ast)}$, since from (i), it would follow that
\begin{itemize}

\item[]  For all $Z$, if $(X\vee Y)\shortTo Z\notin \cbel{\Theta_1}\cap \cbel{\Theta_2}$, then $Y\shortTo Z\in \cbel{\Theta_1}\cap \cbel{\Theta_2}$

\end{itemize}
which is inconsistent with (ii) and (iii).
\end{myproof}

\vspace{1em}

\ModusTollens*

\begin{myproof}
Semantically, the claim translates into: If $\min(\preccurlyeq_{\Psi}, W)\subseteq \mods{A}$, then $\min(\preccurlyeq_{\Psi {\ast_{\mathrm{BG}}} A\shortTo B}, W)\subseteq\mods{A\wedge B}$. So assume $\min(\preccurlyeq_{\Psi}, W)\subseteq \mods{A}$. We will first show that 
\begin{itemize}
\item[] $\min(\preccurlyeq_{\Psi {\contract_{\mathrm{BG}}} A\shortTo \neg B}, W)\subseteq \mods{A}$. 
\end{itemize}
To do so, we  assume that $y\in\mods{\neg A}$ and show that there exists $x$ such that $x\prec_{\Psi {\contract_{\mathrm{BG}}} A\shortTo \neg B}y$ and hence that $y\notin\min(\preccurlyeq_{\Psi {\contract_{\mathrm{BG}}} A\shortTo \neg B}, W)$. From $y\in\mods{\neg A}$ and $\min(\preccurlyeq_{\Psi}, W)\subseteq \mods{A}$, we have $y\notin\min(\preccurlyeq_{\Psi}, W)$. Let $x\in \min(\preccurlyeq_{\Psi}, W)$ and so, since $y\notin\min(\preccurlyeq_{\Psi}, W)$, $x\prec_{\Psi} y$. Since $\min(\preccurlyeq_{\Psi}, W)\subseteq \mods{A}$, we have $x\in\mods{A}$. So we consider two cases:  
\begin{itemize}

\item[{\bf (a)}] $x\in\mods{A\wedge B}$: Then $x\in \min(\preccurlyeq_{\Psi}, \mods{A\wedge B})$. Since $ \min(\preccurlyeq_{\Psi}, \mods{A})=\min(\preccurlyeq_{\Psi}, W)$ and $y\in\mods{\neg A}$,  there is no $z\in\min(\preccurlyeq_{\Psi}, \mods{A})$ such that $y\preccurlyeq_{\Psi} z$. Hence we can apply (2)(b) of Definition \ref{BGcontract}  to recover $x\prec_{\Psi {\contract_{\mathrm{BG}}} A\shortTo \neg B}y$, as required.

\item[{\bf (b)}] $x\in\mods{A\wedge \neg B}$: Then $x\notin \min(\preccurlyeq_{\Psi}, \mods{A\wedge B})$. Furthermore, since $y\in\mods{\neg A}$, $y\notin \min(\preccurlyeq_{\Psi}, \mods{A\wedge B})$. Hence we can apply (1) of Definition \ref{BGcontract}   to recover $x\preccurlyeq_{\Psi{\contract_{\mathrm{BG}}} A\shortTo \neg B} y$ iff $x\preccurlyeq_{\Psi} y$. Since $x\prec_{\Psi} y$, we have $x\prec_{\Psi {\contract_{\mathrm{BG}}} A\shortTo \neg B}y$, as required.

\end{itemize}
We now show that, given what we have just established, 
\begin{itemize}
\item[] $\min(\preccurlyeq_{(\Psi {\contract_{\mathrm{BG}}} A\shortTo \neg B) {+_{\mathrm{BG}}}A\shortTo B}, W)\subseteq\mods{A}$. 
\end{itemize}
We follow the same kind of strategy as above: we assume that $y\in\mods{\neg A}$ and show that there exists $x$ such that $x \prec_{(\Psi {\contract_{\mathrm{BG}}} A\shortTo \neg B) {+_{\mathrm{BG}}}A\shortTo B}y$ and hence that $y\notin\min(\preccurlyeq_{(\Psi {\contract_{\mathrm{BG}}} A\shortTo \neg B) {+_{\mathrm{BG}}}A\shortTo B} , W)$. From $y\in\mods{\neg A}$ and $\min(\preccurlyeq_{\Psi {\contract_{\mathrm{BG}}} A\shortTo \neg B}, W)\subseteq \mods{A}$, we have $y\notin \min(\preccurlyeq_{\Psi {\contract_{\mathrm{BG}}} A\shortTo \neg B}, W)$. Let $x\in \min(\preccurlyeq_{\Psi {\contract_{\mathrm{BG}}} A\shortTo \neg B}, W)$ and so, since $y\notin\min(\preccurlyeq_{\Psi {\contract_{\mathrm{BG}}} A\shortTo \neg B}, W)$, $x\prec_{\Psi {\contract_{\mathrm{BG}}} A\shortTo \neg B} y$. Since $\min(\preccurlyeq_{\Psi {\contract_{\mathrm{BG}}} A\shortTo \neg B}, W)\subseteq \mods{A}$, we have $x\in\mods{A}$. So we consider two cases: 
\begin{itemize}

\item[{\bf (a)}] $x\in\mods{A\wedge B}$: Then $x\notin \min(\preccurlyeq_{\Psi {\contract_{\mathrm{BG}}} A\shortTo \neg B}, \mods{A\wedge \neg B})$, and so we can infer, from condition (1) of Definition \ref{BGexpand}, that $x\preccurlyeq_{(\Psi {\contract_{\mathrm{BG}}} A\shortTo \neg B) {+_{\mathrm{BG}}}A\shortTo B} y$ iff $x\preccurlyeq_{\Psi {\contract_{\mathrm{BG}}} A\shortTo \neg B} y$. But we know that $x\prec_{\Psi {\contract_{\mathrm{BG}}} A\shortTo \neg B} y$. So $x \prec_{(\Psi {\contract_{\mathrm{BG}}} A\shortTo \neg B) {+_{\mathrm{BG}}}A\shortTo B} y$, as required.

\item[{\bf (b)}] $x\in\mods{A\wedge \neg B}$: Since $y\in\mods{\neg A}$, $y\notin\min(\preccurlyeq_{\Psi {\contract_{\mathrm{BG}}} A\shortTo \neg B}, \mods{A\wedge \neg B})$. Furthermore, we know that $x\prec_{\Psi {\contract_{\mathrm{BG}}} A\shortTo \neg B} y$. Hence, by condition (2)(b) of Definition \ref{BGexpand}, we have $x \prec_{(\Psi {\contract_{\mathrm{BG}}} A\shortTo \neg B) {+_{\mathrm{BG}}}A\shortTo B} y$, as required.

\end{itemize}
From $\min(\preccurlyeq_{(\Psi {\contract_{\mathrm{BG}}} A\shortTo \neg B) {+_{\mathrm{BG}}}A\shortTo B}, W)\subseteq\mods{A}$, it follows that $\min(\preccurlyeq_{\Psi {\ast_{\mathrm{BG}}} A\shortTo B}, W)\subseteq\mods{A}$. In other words: $A\in\bel{\Psi {\ast_{\mathrm{BG}}} A\shortTo B}$. By Success, which ${\ast_{\mathrm{BG}}}$ satisfies \cite[Proposition 4]{DBLP:conf/aaai/BoutilierG93} , $A\shortTo B\in\cbel{\Psi {\ast_{\mathrm{BG}}} A\shortTo B}$. By the Ramsey Test, $B\in\bel{ \Psi {\ast_{\mathrm{BG}}} A\shortTo B}$. So $A\wedge B\in\bel{\Psi {\ast_{\mathrm{BG}}} A\shortTo B}$, i.e.~$\min(\preccurlyeq_{\Psi {\ast_{\mathrm{BG}}} A\shortTo B}, W)\subseteq\mods{A\wedge B}$, as required.
\end{myproof}

\vspace{1em}

%% The file kr.bst is a bibliography style file for BibTeX 0.99c
\bibliographystyle{kr}
\bibliography{RBC}

%\vfill
%
%\pagebreak

\end{document}